\documentclass[simfonts,keywords,hyperref]{preprint}

\usepackage{multirow}
\usepackage{amsmath,amssymb,amsfonts}
\usepackage{booktabs}
\usepackage{algorithm}
\usepackage{algorithmicx}
\usepackage{algpseudocode}
\usepackage{listings}
\usepackage{fvextra}
\usepackage{pdflscape}
\usepackage{longtable}
\usepackage{array}
\fvset{breaklines=true, breakanywhere=false, fontsize=\small}
\providecommand{\tightlist}{\setlength{\itemsep}{0pt}\setlength{\parskip}{0pt}}

\title[AIGENIE Tutorial]{The Ultimate Tutorial for AI-driven Scale Development in Generative Psychometrics: Releasing AIGENIE from its Bottle}

\author{Lara L. Russell-Lasalandra}
\affiliation{Department of Psychology, University of Virginia, Charlottesville, VA 22903, USA}
\email{LLR7CB@Virginia.Edu}

\author{Hudson F. Golino}
\affiliation{Department of Psychology, University of Virginia, Charlottesville, VA 22903, USA}
\email{hfg9s@virginia.edu}

\author{Luis Eduardo Garrido}
\affiliation{Department of Psychology, Pontificia Universidad Cat\'olica Madre y Maestra, Dominican Republic}
\email{luisgarrido@pucmm.edu.do}

\author{Alexander P. Christensen}
\affiliation{Department of Psychology, Vanderbilt University, USA}
\email{alexander.christensen@vanderbilt.edu}


\begin{document}

\begin{abstract}
Psychological scale development has traditionally required extensive expert involvement, iterative revision, and large-scale pilot testing before psychometric evaluation can begin. The \texttt{AIGENIE} R package implements the AI-GENIE framework (Automatic Item Generation and Validation with Network-Integrated Evaluation), which integrates large language model (LLM) text generation with network psychometric methods to automate the early stages of this process. The package generates candidate item pools using LLMs, transforms them into high-dimensional embeddings, and applies a multi-step reduction pipeline --- Exploratory Graph Analysis (EGA), Unique Variable Analysis (UVA), and bootstrap EGA --- to produce structurally validated item pools entirely \textit{in silico}. This tutorial introduces the package across eight parts: installation and setup, text generation, embeddings, item generation, the full AI-GENIE pipeline, the GENIE pipeline for researcher-supplied items, advanced prompt engineering, and fully local operation. Two running examples illustrate the package's use: the Big Five personality model (a well-established construct) and AI Anxiety (an emerging construct). The package supports multiple LLM providers (OpenAI, Anthropic, Groq, HuggingFace, and local models), offers a fully offline mode with no external API calls, and provides the \texttt{GENIE()} function for researchers who wish to apply the psychometric reduction pipeline to existing item pools regardless of their origin. The \texttt{AIGENIE} package is freely available on CRAN at \url{https://CRAN.R-project.org/package=AIGENIE}.
\end{abstract}

\maketitle

\section*{Author Note}

\noindent Lara L. Russell-Lasalandra \url{https://orcid.org/0009-0000-3014-1937}

\noindent Hudson F. Golino \url{https://orcid.org/0000-0002-1601-1447}

\noindent Luis Eduardo Garrido \url{https://orcid.org/0000-0001-8932-6063}

\noindent Alexander P. Christensen \url{https://orcid.org/0000-0002-9798-7037}

Luis Eduardo Garrido is supported by Grant 2024-3-5A5-0993 from the Fondo Nacional de Innovaci\'on, Desarrollo Cient\'ifico y Tecnol\'ogico (FONDOCYT) of the Dominican Republic. We have no conflicts of interest to disclose.

Correspondence concerning this article should be addressed to Hudson F. Golino, Department of Psychology, University of Virginia, 485 McCormick Road, Room 216B, Charlottesville, VA 22904, United States. Email: \href{mailto:hfg9s@virginia.edu}{hfg9s@virginia.edu}

\section*{Introduction}\label{introduction}

Large language models (LLMs) are an invaluable tool for scale development. For instance, LLMs can generate extremely high-quality text \parencite{tengler2025exploring,chakrabarty2026}. Using LLMs for text generation, however, is only one piece of the puzzle. Encoder LLMs are also extremely powerful tools because they translate the context and meaning of human language into a numeric, computer-readable vector \parencite{vaswani2017attention,asudani2023impact,tao2025}. Given the sheer cost of traditional item development \parencites[see][]{boateng2018best,clark2019constructing}, researchers have begun using LLMs to write scale items. A growing body of literature has shown that LLM-generated items can meet the same quality benchmarks expected of expert-authored items \parencite{hommel2022transformer,gotz2024let,shin2025examining,keane2026using}, and in some cases even surpass them \parencite{martin2025}.

The present paper details how scale developers and methodologists can harness text generation, embedding models, or both to accelerate the scale development process in an R package called \texttt{AIGENIE} \parencite[short for \emph{Automatic Item Generation and Validation with Network-Integrated Evaluation};][]{russelllasalandra2026aigenie}. \texttt{AIGENIE} is free to use for non-commercial purposes and completely open-source. The package is available on CRAN at \url{https://CRAN.R-project.org/package=AIGENIE} \parencite{aigenie2026software}.

This R package serves an emerging research domain termed \emph{Generative Psychometrics} \parencite{russelllasalandra2026aigenie}, in which language itself is treated as a resource that can be evaluated and assessed algorithmically. Research in adjacent literatures has established that item text alone carries much of what a psychometrician needs from an item. For instance, transformer models can recover the construct an item was written to measure more accurately than human raters \parencite{fyffe2024transforming}, and can predict the correlation between two items from their wording \parencite{hernandez2023aiip}. Additionally, LLMs are becoming credible instruments for psychological assessment \parencite{brickman2025llm}. That is, certain psychometric item analyses are now possible before recruiting any participants. For example, \textcite{garrido2026estimating} use LLM-generated item pools to compare Principal Component Analysis \parencite[PCA;][]{pearson1901lines} against network-based methods for recovering dimensional structure without surveying any human sample. The in-silico nature of this discovery and others within Generative Psychometrics hinges on relatively recent advancements (i.e., the self-attention mechanism \parencite{vaswani2017attention}) in \emph{embedding} technology. Embeddings are high-dimensional numerical vectors that represent the meaning and context of human language. \texttt{AIGENIE} can use such embeddings to quantitatively evaluate items \emph{prior to any human data collection}. Structural validation of entire scales is possible \emph{in silico} as the structure estimated using item embeddings matches the structure estimated in representative samples of US adults \parencite{russelllasalandra2026aigenie}. This approach substantially reduces resource barriers that have long characterized measurement development.

AI-GENIE evaluates these embeddings using \emph{network psychometrics}, an approach that represents a construct not as a latent variable causing its indicators, but as a system of items whose interrelations are modeled directly with groups of items forming clusters. Within a psychometric network, each item is a node, and an edge between two nodes represents the association between that pair of items after controlling for every other item in the set. A ``strong'' edge between nodes means items are closely related, whereas the absence of an edge indicates two items that share nothing beyond what other items already explain. In AI-GENIE, items' embedding vectors (as opposed to item responses) are used to build the network. This substitution can be done because item wording carries the semantic information reflected in response patterns. Evidence indicates that quality embedding vectors can adequately stand in for item responses in this context \parencite{fang2022construct,hommel2025infer,kambeitz2025structure,kilmen2025shortening}. Furthermore, items that assess the same facet of a latent construct tend to form densely interconnected subsets of the network, known as \emph{communities}. Why? Say, for example, that five items in a depression inventory assess feelings of self-doubt while five items assess disordered sleeping habits. All of the sleep items would have a lot of semantic and conceptual overlap with one another, and the same would be true of the self-doubt items. Thus, the embedding vectors, which capture item context, would reflect this difference between item types. In this ten-item network, we would therefore expect to see two communities. Exploratory Graph Analysis \parencite[EGA;][]{golino2017exploratory} estimates this network and then applies a community detection algorithm to identify those subsets. The resulting communities are interpreted as dimensions (or latent constructs/facets), such that the number of communities EGA recovers is analogous to the number of factors retained in exploratory factor analysis, and each item's community membership is analogous to its pattern of loadings. EGA has been shown to recover dimensional structure at least as accurately as traditional factor-analytic retention methods \parencite{golino2020investigating}.

The structure EGA recovers can be compared directly against the structure the researcher intended. Our package compares the intended, theoretical communities or clusters of items with the derived communities using \emph{Normalized Mutual Information} \parencite[NMI;][]{danon2005comparing}. \emph{NMI} is measured on a scale from 0-1 (here, scaled by a factor of 100 to be represented as a percentage), in which a score of 0\% indicates complete dissimilarity between the known and empirically derived communities and a score of 100\% represents a perfect match. NMI can loosely be thought of as an ``accuracy'' metric, where higher scores mean better community detection. Note that the objects returned by the package store NMI as a proportion (e.g., \texttt{0.955}), whereas the summary printed to the console reports it on a 0-100 scale (e.g., \texttt{95.55}).

In our previous ten-item example, the researcher would expect to see two communities with five items each, and the items within each community should represent the sleeping and self-doubt facets. If, say, a sleeping item was erroneously clustering within the self-doubt community, this unexpected item community assignment could be signaling that this misaligned sleeping item is of poor quality.

This evidence becomes stronger if the misalignment repeats across several hundred networks built by bootstrapping the original embedding matrix. An item's propensity to cluster consistently within the same community across those networks is its \emph{structural stability}. \textcite{Haladyna2013} frame item validation as an interpretive argument. They state a chain of sixteen questions, each answered by a different kind of evidence, in which failure to satisfy any one is a weakness in the chain supporting validity. AI-GENIE answers one of them: whether the internal structure of the test content has been studied. AI-GENIE checks both the agreement between the communities EGA recovers and the intended structure the researcher specified and how robustly these recovered community assignments hold. This \textbf{structural validation} describes the aspect of validity that AI-GENIE assesses. Other forms of validation (e.g., content validation) are \emph{not} covered by AI-GENIE.

\subsection*{The AIGENIE Reduction Pipeline}\label{the-aigenie-reduction-pipeline}

\begin{figure}[!htbp]
    \centering
    \includegraphics[width=0.95\linewidth]{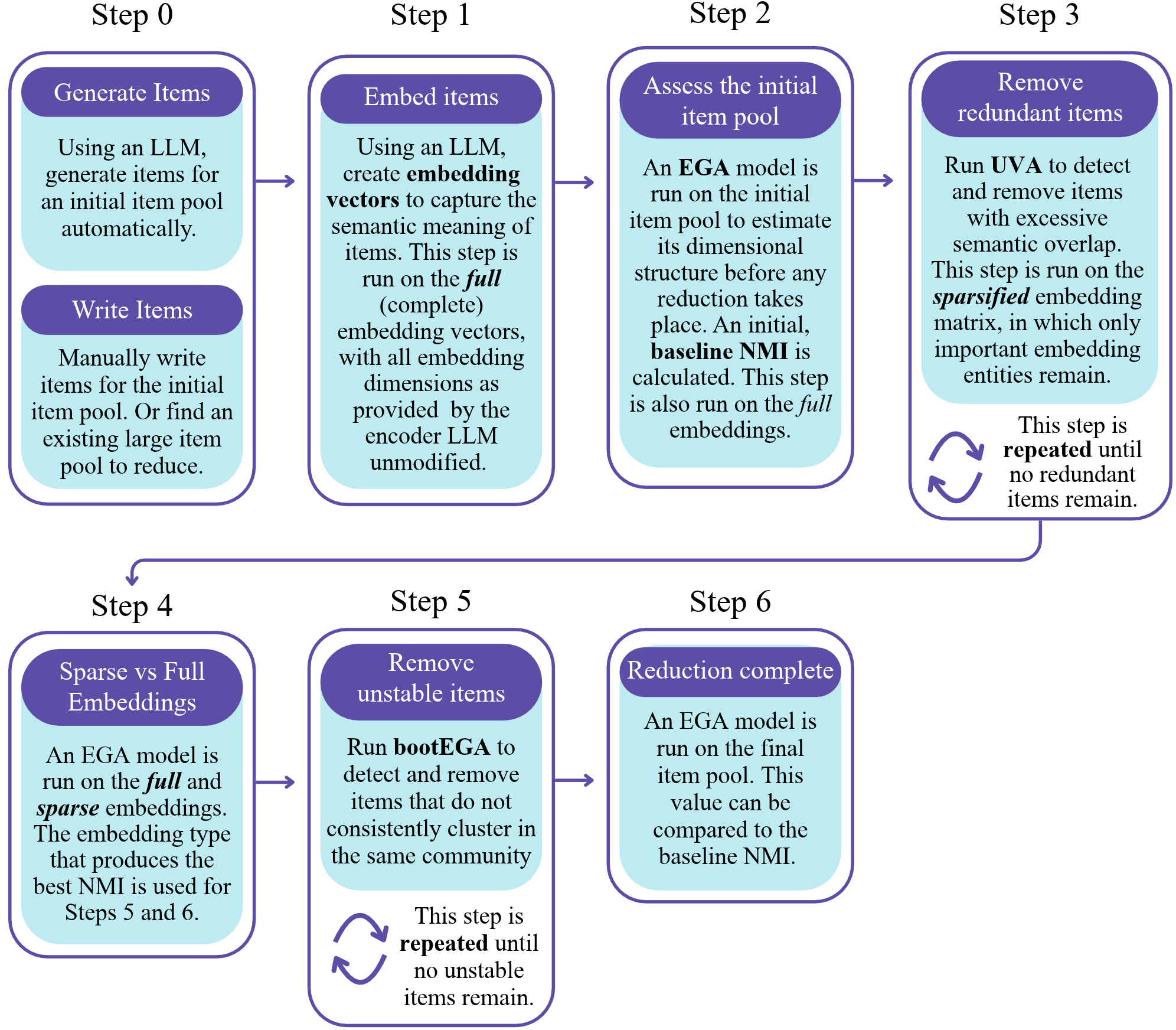}
    \caption{\textit{AI-GENIE Reduction Pipeline.} This figure illustrates the entire AI-GENIE pipeline. Step 0, which precedes item reduction, is obtaining the initial item pool, either by generating items with an LLM or by writing them manually. Step 1 obtains the item embeddings using an encoder LLM. Steps 2 through 6 apply network psychometric techniques to reduce the item pool algorithmically.}
    \label{fig:fig-pipeline}
\end{figure}

The AIGENIE methodology implemented within this R package combines (optional) item generation with network psychometric techniques for structural validation. AI-GENIE uses LLMs to generate large candidate item pools (or uses an existing pool of items authored by humans), embeds them as high-dimensional vectors, and then applies a multi-step psychometric pipeline to identify and remove redundant or unstable items. This pipeline includes Exploratory Graph Analysis \parencite[EGA;][]{golino2017exploratory}, Unique Variable Analysis \parencite[UVA;][]{christensen2023unique}, and bootstrap EGA \parencite{christensen2021estimating}. The resulting item pool is a more concise, structurally validated set of items ready for empirical testing. The seven steps (numbered 0 through 6) of the AI-GENIE pipeline are as follows (see Figure \ref{fig:fig-pipeline}):

\begin{enumerate}
\def\labelenumi{\arabic{enumi}.}
\setcounter{enumi}{-1}
\item
  \textbf{Generate or write your initial item pool.} The item reduction process must begin with a sizable item pool. These initial items can either be generated seamlessly within the package using an LLM or directly supplied by the user.
\item
  \textbf{Embed items.} Each item is transformed into a high-dimensional numeric vector (an \emph{embedding}) using an encoder LLM. Encoder LLMs convert human-readable text into these numeric embedding vectors. In this case, an encoder LLM will translate the items' text into a numeric format for quantitative analyses, and these analyses inform item deletion. More powerful encoder LLMs will produce higher quality embeddings (i.e., better ``translations'' of the meaning and context of human language); the quality of the AI-GENIE reduction relies foremost on the quality of its inputs. OpenAI's \emph{text-embedding-3-small} model has performed well in our simulations \parencite{garrido2026estimating}.
\item
  \textbf{Assess the initial item pool.} An EGA model is run on the initial item pool to estimate its dimensional structure before any reduction takes place. The detected communities (clusters of items) are compared to the known, intended structure by computing an NMI score. This initial NMI establishes a baseline measure of structural validity. This step is run on the \emph{full}, complete embedding matrix; that is, numbers in all embedding dimensions are kept without alteration, just as produced by the encoder LLM.
\item
  \textbf{Remove redundant items.} UVA is used iteratively to detect and remove items with excessive semantic overlap. UVA identifies redundant item pairs or sets based on weighted topological overlap \parencite[wTO;][]{zhang2005general} within the network, retaining only the most unique representative from each redundant cluster. This step repeats until no further redundancies are found. This step is run on a \emph{sparsified} version of the matrix (in which only the most informative embedding dimensions are retained).
\item
  \textbf{Select sparse or full embeddings.} An EGA model is run on the \emph{full} embedding matrix and the \emph{sparsified} version. The embedding type that corresponds to the EGA network with the better NMI is retained for all subsequent steps.
\item
  \textbf{Find the most stable items.} BootEGA is used to assess the structural stability of each item. In this step, several hundred new embedding matrices are generated by drawing from a multivariate normal distribution parameterized by the original embedding matrix, and then EGA is applied to each one. If an item is consistently assigned to the same dimension across the resamples, it is considered stable; items that frequently shift between communities are considered unstable and removed. This step repeats until all remaining items demonstrate high stability.
\item
  \textbf{Final pool is ready for review.} A final EGA model is run on the reduced item pool. The final NMI is calculated to assess the quality of said detection after item pool reduction. This NMI measure can be compared to that of the baseline from Step 2.
\end{enumerate}

The efficacy of AI-GENIE has been demonstrated through large-scale Monte Carlo simulations across multiple LLMs, showing consistent improvements in structural validity across all conditions \parencite{russelllasalandra2026aigenie}. A follow-up simulation study further demonstrated that the prompt engineering strategy of adaptive prompting \parencite{lightman2023letsverifystepstep} can meaningfully shape the quality of AI-generated item pools, with effects that scale with model capability \parencite{russelllasalandra2025prompt}. The \texttt{AIGENIE} R package implements adaptive prompting by default.

\subsection*{Overview of the Tutorial}\label{overview-of-the-tutorial}

This tutorial provides a comprehensive, step-by-step guide to using the \texttt{AIGENIE} R package. We organize the tutorial into eight parts that progressively introduce the package's capabilities:

\begin{itemize}
\item
  \textbf{Part 1} covers installation and setup, Python environment configuration, and an overview of package usage fundamentals (e.g., management of API-based LLMs).
\item
  \textbf{Part 2} introduces the text generation capabilities of the package through the \texttt{chat()} and \texttt{local\_\allowbreak{}chat()} functions, which allow users to interact with LLMs directly from R.
\item
  \textbf{Part 3} covers the embedding functionality, explaining the different embedding providers available (OpenAI, Jina AI, HuggingFace, and local models) and how to generate embeddings for item pools.
\item
  \textbf{Part 4} demonstrates standalone item generation using the \texttt{AIGENIE()} function's \texttt{items.\allowbreak{}only} mode, which generates items without running the full psychometric pipeline.
\item
  \textbf{Part 5} walks through the complete AI-GENIE pipeline from start to finish, first with a simple Big Five personality example and then with a more extended AI Anxiety \parencite[AIA;][]{wang2022development} demonstration.
\item
  \textbf{Part 6} introduces the \texttt{GENIE()} function, which applies only the psychometric evaluation and reduction pipeline to user-supplied item pools --- a critical capability for researchers who already have items and want to assess their structural validity.
\item
  \textbf{Part 7} covers advanced prompt engineering options, drawing on the findings from the prompt engineering simulation study to show users how to customize prompts, system roles, and adaptive generation settings for optimal results.
\item
  \textbf{Part 8} demonstrates fully local operation via \texttt{local\_\allowbreak{}AIGENIE()}, \texttt{local\_\allowbreak{}GENIE()}, and \texttt{local\_\allowbreak{}chat()}, in which item generation and embedding run entirely on the user's own machine with no external API calls.
\end{itemize}

\texttt{AIGENIE} provides a unified, open-source toolkit that brings together LLM-based text generation/embedding and network psychometric evaluation. By automating the most labor-intensive stages of scale development (i.e., item writing), the package lowers the resource barriers that have historically limited who can develop high-quality measurement instruments and how quickly they can do so. \texttt{AIGENIE} offers a flexible entry point into the Generative Psychometrics paradigm. The present paper is intended to make each step of that process as accessible as possible, regardless of the reader's prior experience with LLMs or network analysis.

\section{Part 1: Package Setup and Important Notes}\label{part1}

Setting up the \texttt{AIGENIE} package requires several steps, chief among them configuring the Python backend. Optionally, researchers can also install support for fully local models. The sections below walk through each requirement in order, with platform-specific instructions for macOS/Linux and Windows.

Although \texttt{AIGENIE} is an R package, it relies on a Python backend because the libraries that call LLM APIs and run local models are developed primarily in the Python ecosystem. The package uses \texttt{reticulate} \parencite{ushey2023reticulate} to call those libraries directly from R, so users never write or see Python code, and \texttt{uv} \parencite{astral2024uv} to manage the Python environment in which they live. The \texttt{uv} package is a Rust-based manager that installs and produces isolated environments that do not interfere with any other Python installation on the system.

\subsection{Install R and RStudio}\label{install-r-and-rstudio}

We recommend installing the latest version of both \href{https://cran.r-project.org/}{R} and \href{https://posit.co/download/rstudio-desktop/}{RStudio}. Installation of both is straightforward; if you need step-by-step guidance, follow these \href{https://rstudio-education.github.io/hopr/starting.html}{instructions}.

\subsection{Install Command Line Tools (macOS/Linux Only)}\label{install-command-line-tools-macoslinux-only}

macOS users need Apple's Command Line Tools to compile certain dependencies. Open the \textbf{Terminal} app and run:

\begin{lstlisting}[language=bash]
xcode-select --install
\end{lstlisting}

A dialog will appear prompting you to install the tools. Follow the on-screen instructions. Linux users can skip this step, as the necessary build tools are typically pre-installed or available through the system package manager.

\subsection{Install the AIGENIE Package}\label{install-the-aigenie-package}

Next, install the \texttt{AIGENIE} package. The package is available on CRAN:

\begin{Verbatim}
# Install AIGENIE from CRAN
install.packages("AIGENIE")
\end{Verbatim}

\subsection{\texorpdfstring{Install the \texttt{uv} Python Environment Manager}{Install the uv Python Environment Manager}}\label{uv-install}

\texttt{AIGENIE} communicates with LLMs through Python, accessed seamlessly from R via the \texttt{reticulate} and \texttt{uv} packages. You do \textbf{not} need to know Python to use \texttt{AIGENIE} -- the packages handle all Python interactions in the background. However, you do need to ensure that \texttt{uv} is installed properly on your machine.

\subsubsection{macOS / Linux}\label{macos-linux}

Open your \textbf{Terminal} and run:

\begin{lstlisting}[language=bash]
curl -LsSf https://astral.sh/uv/install.sh | sh
\end{lstlisting}

If successful, you should see a message ending in ``everything's installed!'' It should \emph{not} say ``permission denied'' anywhere in the output. If you do see a permission error, run the following in your Terminal and then retry the \texttt{curl} command above:

\begin{lstlisting}[language=bash]
echo 'export PATH="$HOME/.local/bin:$PATH"' >> ~/.zshrc
source ~/.zshrc
\end{lstlisting}

\subsubsection{Windows}\label{windows}

Open \textbf{Windows PowerShell} (search for it in your Start menu). First, set the execution policy to allow scripts:

\begin{lstlisting}[language=bash]
Set-ExecutionPolicy RemoteSigned -Scope CurrentUser
\end{lstlisting}

Then install \texttt{uv}:

\begin{lstlisting}[language=bash]
irm https://astral.sh/uv/install.ps1 | iex
\end{lstlisting}

If successful, you should see a message ending in ``everything's installed!'' For further troubleshooting, see the \href{https://docs.astral.sh/uv/getting-started/installation/}{uv installation documentation}.

\emph{After installing \texttt{uv}, save your R script, restart your R session, and restart your computer.} This ensures that your system's PATH is updated and R can locate the \texttt{uv} installation.

\subsection{Verify the Installation}\label{verify-the-installation}

Load the package and confirm that R can detect the \texttt{uv} installation:

\begin{Verbatim}
library(AIGENIE)
check_installation <- python_env_info()
check_installation[["uv_available"]]  # This should be TRUE
\end{Verbatim}

If \texttt{uv\_\allowbreak{}available} returns \texttt{FALSE} despite a successful installation in Section \ref{uv-install}, R may not be able to find \texttt{uv} on your system's PATH. On \textbf{macOS/Linux}, open Terminal and run:

\begin{lstlisting}[language=bash]
PATH="/usr/local/bin:/usr/bin:/bin:/usr/sbin:/sbin:$HOME/.local/bin:$HOME/.cargo/bin:$PATH"
\end{lstlisting}

On \textbf{Windows}, open PowerShell and run (replacing \texttt{\textless{}you\textgreater{}} with your username):

\begin{lstlisting}[language=bash]
[Environment]::SetEnvironmentVariable("PATH", $env:PATH + ";C:\Users\<you>\.local\bin", "User")
\end{lstlisting}

After applying either PATH fix, you \emph{must} fully close and reopen RStudio (not just restart the R session) for the changes to take effect.

\subsection{Python Environment Setup}\label{python-environment-setup}

The first time you call any \texttt{AIGENIE} function that requires Python (e.g., \texttt{chat()}, \texttt{AIGENIE()}, \texttt{GENIE()}), the package will automatically create a dedicated Python virtual environment and install the necessary dependencies. This process typically takes two or three minutes and only needs to happen once.

If you prefer to set up the environment explicitly, or if you need to customize the installation, you can use the following functions:

\begin{Verbatim}
# Standard setup (happens automatically on first use)
ensure_aigenie_python()

# If something goes wrong, reinstall the environment from scratch
reinstall_python_env()
\end{Verbatim}

You can inspect the current Python environment at any time:

\begin{Verbatim}
python_env_info()
\end{Verbatim}

\subsection{Install Local Model Support (Only Required When Using Local Models)}\label{local-setup}

The setup described so far is sufficient for using \texttt{AIGENIE} with LLMs accessed via an Application Programming Interface (API; examples include OpenAI and Anthropic). More detailed information on API use can be found in Section \ref{APIkeys} below. However, a select handful of users may want to run models entirely on their own machine (that is, generating items and embeddings without sending any data to external servers). These users must install additional local model support. This is required for the \texttt{local\_\allowbreak{}AIGENIE()}, \texttt{local\_\allowbreak{}GENIE()}, and \texttt{local\_\allowbreak{}chat()} functions described in Parts \ref{part2} and \ref{part8}.

Running models locally is neither necessary nor recommended for \emph{most} users. Models that run comfortably on consumer hardware (typically 7--13 billion parameters) are considerably less capable than frontier API models; larger models demand a high-end GPU with substantial VRAM, so users looking to run models locally must have a machine with substantial compute. Local operation is best reserved for cases where data privacy requirements prohibit sending item content to external servers. For all other purposes we recommend an API-based provider.

\subsubsection{macOS / Linux}\label{macos-linux-1}

On macOS and Linux, local model support can be installed directly from within R:

\begin{Verbatim}
install_local_llm_support()
\end{Verbatim}

For Apple Silicon Macs, this function automatically compiles \texttt{llama-cpp-python} with Metal acceleration for GPU-accelerated inference. For users with NVIDIA GPUs, additionally run \texttt{install\_\allowbreak{}gpu\_\allowbreak{}support()} to install CUDA-enabled PyTorch for accelerated embedding generation.

\subsubsection{Windows}\label{windows-1}

Windows users need one additional prerequisite before \texttt{install\_\allowbreak{}local\_\allowbreak{}llm\_\allowbreak{}support()} will work: the \textbf{C++ Build Tools} from Microsoft. These are required to compile \texttt{llama-cpp-python}.

\begin{enumerate}
\def\labelenumi{\arabic{enumi}.}
\tightlist
\item
  Download the \href{https://visualstudio.microsoft.com/visual-cpp-build-tools/}{Visual Studio C++ Build Tools}.
\item
  Open the \textbf{Visual Studio Installer}. Next to ``Visual Studio Build Tools 2026,'' click \textbf{Modify}.
\item
  Check \textbf{``Desktop Development with C++''} and ensure the first six options in the right panel are selected.
\item
  Click \textbf{Modify} in the bottom-right corner. This installation is over 1 GB and may take several minutes.
\end{enumerate}

Once the C++ Build Tools are installed, run the following in R:

\begin{Verbatim}
install_local_llm_support()
\end{Verbatim}

\subsubsection{Downloading a Local Model}\label{downloading-a-local-model}

With local support installed, you can download GGUF-format models from HuggingFace using the built-in \texttt{get\_\allowbreak{}local\_\allowbreak{}llm()} function. We recommend \emph{Mistral 7B Instruct} to test your setup, given its relative efficiency for item generation:

\begin{Verbatim}
model_path <- get_local_llm(
  repo_id = "TheBloke/Mistral-7B-Instruct-v0.2-GGUF",
  filename = "mistral-7b-instruct-v0.2.Q4_K_M.gguf"
)

# The function returns the path where the model was saved
print(model_path)
\end{Verbatim}

To verify that local model support is properly configured:

\begin{Verbatim}
check_local_llm_setup(model_path)
\end{Verbatim}

\subsection{Understanding APIs and API Keys}\label{APIkeys}

An Application Programming Interface (API) provides a way for the user's machine to communicate with and request tasks from AI providers' servers. When \texttt{AIGENIE} generates items or computes embeddings using a cloud-based model, it does so by sending a request to the model provider's API. Therefore, the user never interacts with the API directly; the package handles all API communication behind the scenes.

To authenticate these requests, providers require an \textbf{API key.} This key is a unique string of characters that functions as both an ID \emph{and} a password. When \texttt{AIGENIE} sends a request to, say, Groq's servers, it includes your Groq API key so that the server can verify your identity, check that you have permission to use the requested model, complete the requested task (if able), and log the usage for billing purposes.

Much like a password, API keys should \emph{never} be shared, and they should be stored securely. Anyone with your API key can potentially impersonate you, make requests, and add charges to your account. If you suspect that your key has been lost or stolen, you can very easily delete (or revoke) your API key and create a new one. Both \href{https://help.openai.com/en/articles/5112595-best-practices-for-api-key-safety}{OpenAI} and \href{https://docs.cloud.google.com/docs/authentication/api-keys-best-practices}{Google Cloud} have further guidance on best practices for API key usage.

\subsubsection{Supported Providers}\label{supported-providers}

As shown in Table \ref{tab:api-providers}, \texttt{AIGENIE} supports API keys from five providers (OpenAI, Anthropic, Groq, HuggingFace, and Jina).

Users will have to create an account with whichever provider or providers they have chosen; this account will be used to enable users to create an API key as well as track their usage.

Note that API usage is tracked in \textbf{tokens}, where a token is roughly three-quarters of a word \parencite{openai2024tokens}. Providers typically charge separately for input tokens (the prompt you send) and output tokens (the text the model generates), with output tokens priced higher. Additionally, each provider imposes limits on how frequently you can send requests within a given time window. If you exceed these limits, the API will temporarily reject your requests. Cost and rate limits vary widely per provider \emph{and} per model. Users can find detailed and up-to-date pricing and rate limiting information on the providers' websites.

\begin{table}[ht]
\centering
\caption{API providers compatible with the AIGENIE package}
\label{tab:api-providers}
\begin{tabular}{p{0.16\textwidth} p{0.18\textwidth} p{0.30\textwidth} p{0.24\textwidth}}
\toprule
\textbf{Provider} & \textbf{Model Types} & \textbf{Cost} & \textbf{Website} \\
\midrule

\textbf{OpenAI} &
text and embedding &
``Pay as you go'' billing. Prepaid credits are also available. &
\url{https://platform.openai.com} \\

\textbf{Anthropic} &
text only &
Prepaid credits only. &
\url{https://console.anthropic.com} \\

\textbf{Groq} &
text only &
Free base plan with opportunity to upgrade to better rate limits. &
\url{https://console.groq.com} \\

\textbf{HuggingFace} &
text and embedding &
Typically free with costs incurring only in special circumstances. &
\url{https://huggingface.co} \\

\textbf{Jina AI} &
embedding only &
Free for the first 10 million tokens. &
\url{https://jina.ai/embeddings} \\

\bottomrule
\end{tabular}
\end{table}

After obtaining API keys from provider websites, they can be passed directly to \texttt{AIGENIE} functions:

\begin{Verbatim}
# Pass keys directly to functions
results <- AIGENIE(
  ...,
  openai.API = "sk-your-openai-key-here",
  groq.API = "gsk_your-groq-key-here"
)
\end{Verbatim}

\subsection{Discovering Available Models}\label{discovering-available-models}

Model availability varies widely by provider, by account, and over time. In fact, it is very likely that a model named in this paper may since have been retired. Rather than relying on a fixed list, \texttt{list\_\allowbreak{}available\_\allowbreak{}models()} queries the providers directly and reports what the supplied keys can actually reach.

\begin{Verbatim}
# check the number of models currently accessible given API keys
models <- list_available_models(
  openai.API = "your-openai-key",
  groq.API = "your-groq-key",
  anthropic.API = "your-anthropic-key"
)
\end{Verbatim}

The function returns a \texttt{data.\allowbreak{}frame} with five columns: \texttt{provider}, \texttt{model} (the identifier to pass to \texttt{model} or \texttt{embedding.\allowbreak{}model}), \texttt{type} (\texttt{"chat"} or \texttt{"embedding"}), \texttt{display\_\allowbreak{}name}, and \texttt{created}. Rows are sorted by provider, then type, then model name.

The unfiltered result is likely going to be too long. To narrow this list, the \texttt{provider} argument restricts the output to one of \texttt{"openai"}, \texttt{"groq"}, \texttt{"anthropic"}, or \texttt{"jina"}. The \texttt{type} argument restricts it to \texttt{"chat"} or \texttt{"embedding"}.

For example, here is how the user could search for which of OpenAI's embedding models are available:

\begin{Verbatim}
# looking only at OpenAI and embedding models
list_available_models(provider = "openai", # or another provider (e.g. 'groq')
                      type = "embedding", # or 'chat' for text models only
                      openai.API = "your-openai-key") 

#>   provider                  model      type           display_name    created
#> 1   openai text-embedding-3-large embedding text-embedding-3-large 2024-01-22
#> 2   openai text-embedding-3-small embedding text-embedding-3-small 2024-01-22
#> 3   openai text-embedding-ada-002 embedding text-embedding-ada-002 2022-12-16
\end{Verbatim}

\subsection{Item Types and Item Attributes}\label{item-types-and-item-attributes}

To effectively use the \texttt{AIGENIE} package, one must understand \textbf{item types} as well as \textbf{item attributes}. For our purposes, an \emph{item type} describes the items' more general/broad construct or second-order dimension of interest. \emph{Item attributes} are the specific facets, first-order dimensions, or content areas that the generated items should collectively cover.

In the AI-GENIE simulation paper, the Big Five personality traits \parencite[openness to experience, conscientiousness, extraversion, agreeableness, and neuroticism;][]{john1999big} were used as the item types. Each of these five item types contained four attributes (or content areas) that we wanted those items to cover:

\begin{enumerate}
\def\labelenumi{\arabic{enumi}.}
\tightlist
\item
  \emph{Openness to experience:} creative, perceptual, curious, and philosophical.
\item
  \emph{Conscientiousness:} organized, responsible, disciplined, and prudent.
\item
  \emph{Extraversion:} friendly, positive, assertive, and energetic.
\item
  \emph{Agreeableness:} cooperative, compassionate, trustworthy, and humble.
\item
  \emph{Neuroticism:} anxious, depressed, insecure, and emotional.
\end{enumerate}

It is important to note that attributes do \emph{not} need to correspond to formal subscales or subdomains. Additionally, items need \emph{not} necessarily be organized into a formal hierarchical structure. For example, in the original AI-GENIE simulation, the AI Anxiety factor ``learning anxiety'' had the attributes \texttt{c("overwhelmed",\ "inadequacy",\ "intimidated")} simply because they reflect the distinct thematic facets that we thought a comprehensive set of learning anxiety items should span. Including attributes ensures that the LLM does not generate items that cluster around a single, extremely narrow aspect of the construct while neglecting others. The model is instructed to produce items for each attribute, which results in a more content-diverse initial pool more appropriate for reduction.

Thus, \texttt{AIGENIE} \emph{requires} that each item type have \emph{at least} two attributes. We recommend three to six, though, in theory, more are possible. The item types need \emph{not} have the same number of attributes (e.g., it would have been entirely plausible to define, say, five attributes for the neuroticism personality trait, three for extraversion, and four for the rest of them). Additionally, the item types and attributes can be more than one word. We recommend no more than four words are used to name the item types/attributes, though, again, more are possible in theory.

\subsubsection{How Types and Attributes are Used in the Package}\label{how-types-and-attributes-are-used-in-the-package}

The most important functions in the \texttt{AIGENIE} package are the \texttt{AIGENIE()} and the \texttt{GENIE()} functions.\footnote{\texttt{local\_\allowbreak{}AIGENIE()} and \texttt{local\_\allowbreak{}GENIE()} are sister functions that operate almost identically, with the exception of how LLMs are called to generate text and/or embeddings. In this section, anything said of the \texttt{AIGENIE()} or \texttt{GENIE()} functions is also true of their corresponding local versions.} Both use the concept of item types and attributes in different ways.

In the \texttt{AIGENIE()} function, items are generated, embedded, and subjected to the full reduction pipeline. The single most important parameter in the \texttt{AIGENIE()} function is \texttt{item.\allowbreak{}attributes}. This parameter is a named list in which each element represents an item type (e.g., one of the Big Five personality traits), and the character vector within each element specifies its attributes. The model will be instructed to generate items that target each of these attributes, producing items across the full breadth of each item type. For example:

\begin{Verbatim}
big5_attributes <- list(
  openness = c("creative", "perceptual", "curious", "philosophical"),
  conscientiousness = c("organized", "responsible", "disciplined", "prudent"),
  extraversion = c("friendly", "positive", "assertive", "energetic"),
  agreeableness = c("cooperative", "compassionate", "trustworthy", "humble"),
  neuroticism = c("anxious", "depressed", "insecure", "emotional")
)
\end{Verbatim}

In the \texttt{GENIE()} function, an existing pool of items runs through the full reduction pipeline. Thus, items must have already been written or generated. These items should be situated within a \texttt{data\ frame} with exactly four columns named \texttt{ID}, \texttt{statement}, \texttt{type}, and \texttt{attribute}. The \texttt{ID} column serves as a simple item identifier (most commonly a number), the \texttt{statement} column contains the actual content of the item, while the \texttt{type} and \texttt{attribute} columns are self-explanatory. Here is how one might construct such a \texttt{data\ frame} for use in \texttt{GENIE()}:

\begin{Verbatim}
# first, draft the item content, keeping in mind its type and attribute
# for the sake of example, only ONE item is given per attribute,
# but many more would be needed in a real run
item_content <- c(
  # openness to experience item type 
  "I can easily think up never-before-used solutions.", # creative 
  "I take note of minor details in my surroundings.", # perceptual 
  "I will readily try new foods, even if completely foreign to me.", # curious
  "I enjoy pondering questions that have implications for humanity.", # philosophical
  ..., # plus many, many more items that target one of these four attributes
  
  # conscientiousness item type
  "I almost never leave something out of its place.", # organized
  "I take special care of nearly everything in my possession.", # responsible
  "When I can set long-term goals, I almost always follow through.", # disciplined 
  "I tend to plan for worst-case scenarios whenever possible.", # prudent
  ..., # plus many, many more items that target one of these four attributes
  
  # extraversion item type 
  "I am neighborly, even to acquaintances and new people.", # friendly
  "I try to look on the bright side of tough situations.", # positive
  "I enjoy taking charge and being the leader in a group.", # assertive 
  "I am usually the life of the party.", # energetic 
  ..., # plus many, many more items that target one of these four attributes
  
  # agreeableness item type 
  "In a group setting, I am usually the one to lend a helping hand.", # cooperative
  "I can easily sympathize with others, even strangers.", # compassionate
  "I am known as the person who always keeps their promises.", # trustworthy
  "I don't enjoy a lot of praise, even if I achieved something challenging.", #humble
  ..., # plus many, many more items that target one of these four attributes
  
  # neuroticism item type
  "I tend to be nervous, even for mundane tasks.", # anxious
  "I struggle to find the motivation to get up in the morning.", # depressed
  "I get jealous very easily.", # insecure 
  "My mood tends to shift on a dime.", # emotional 
  ... # plus many, many more items that target one of these four attributes
)


# create an object that specifies the item types, as per `item_content`
big5_item_type <- c(
  "openness", "openness", ..., "openness", 
  "conscientiousness", "conscientiousness", ..., "conscientiousness", 
  "extraversion", "extraversion", ..., "extraversion",
  "agreeableness", "agreeableness", ..., "agreeableness", 
  "neuroticism", "neuroticism", ..., "neuroticism"
)


# create an object that specifies the item attributes, as per `item_content`
big5_item_attributes <- c(
  "creative", "perceptual", "curious", "philosophical", ..., 
  "organized", "responsible", "disciplined", "prudent", ..., 
  "friendly", "positive", "assertive", "energetic", ..., 
  "cooperative", "compassionate", "trustworthy", "humble", ..., 
  "anxious", "depressed", "insecure", "emotional", ...
)

# create an ID variable to label the items 
big5_item_ID <- as.factor(1:length(item_content))


# assemble everything in a data frame
big5_item_df <- data.frame(
  ID = big5_item_ID, 
  statement = item_content, 
  type = big5_item_type,
  attribute = big5_item_attributes
)
\end{Verbatim}

Of course, reading in a CSV file would be easier than manually building the \texttt{data\ frame} using R code. The above code merely illustrates the expected formatting more explicitly. For a meaningful reduction, we recommend at least 15 items per attribute.

\subsubsection{\texorpdfstring{Item Types, Attributes, and the \texttt{all.\allowbreak{}together} Flag}{Item Types, Attributes, and the all.together Flag}}\label{item-types-attributes-and-the-all.together-flag}

By default, the reduction pipeline processes each item type \emph{independently.} In our example, the openness items would run through the reduction pipeline first, followed independently by conscientiousness, extraversion, agreeableness, and neuroticism. Thus, with \emph{five} item types, the default behavior produces \emph{five} independent pipeline runs and \emph{five} corresponding reduction results.

Within each type-level analysis, NMI is computed by comparing the communities identified by EGA with the items' known attributes. Item type cannot serve as the reference partition because all items entering a given type-level pipeline share the same type.

Setting \texttt{all.\allowbreak{}together\ =\ TRUE} changes this behavior fundamentally. Instead of running separate reduction pipelines for each item type, AIGENIE pools all items and runs the full reduction pipeline on the \emph{combined} item pool. Consequently, UVA can identify redundancy across item types, and EGA, bootEGA, and subsequent reduction steps operate on a single network containing all items.

Internally, AIGENIE temporarily transforms the item metadata so that the entire pool can be processed as one item type. Each item's reference attribute is represented by the combination of its original type and attribute. Thus, for example, an openness item with the attribute \texttt{"creative"} is represented separately from an item belonging to another type that might have the same attribute label. Conceptually, these reference groups correspond to labels such as \texttt{"openness\ creative"} and \texttt{"neuroticism\ anxious"}.

NMI during the pooled analysis is therefore computed by comparing the EGA communities with this type-by-attribute reference structure.

Accordingly, in our five-type example, \texttt{all.\allowbreak{}together\ =\ TRUE} produces a single pooled reduction pipeline and a single reduction result, rather than five independent type-level analyses.

The modification of type and attribute used for the pooled analysis is internal and temporary. After the pipeline has completed, AIGENIE restores each retained item's original type and attribute labels while preserving its estimated \texttt{EGA\_\allowbreak{}comcommunity} assignment. Thus, the returned \texttt{final\_\allowbreak{}items}, and \texttt{initial\_\allowbreak{}items} when requested, retain the original item metadata.

\section{Part 2: Text Generation}\label{part2}

Before diving into the full psychometric pipeline, it is helpful to understand how the package handles direct interaction with LLMs. The \texttt{chat()} and \texttt{local\_\allowbreak{}chat()} functions provide a straightforward interface for sending prompts to language models and receiving responses. These functions are useful for exploratory work, prompt testing, or any task requiring LLM-generated text.

\subsection{Generating Text via API}\label{generating-text-via-api}

The \texttt{chat()} function supports all API-based providers with text generation models (i.e., OpenAI, Anthropic, Groq, and HuggingFace):

\begin{Verbatim}
# Basic usage with OpenAI
response <- chat(
  prompts = "Generate 5 items measuring conscientiousness for a personality scale.",
  model = "OpenAI/gpt-5.4",
  openai.API = "your-key"
)

# View the response
print(response)
\end{Verbatim}

The function returns a data frame with columns for the repetition index, the original prompt, and the model's response. This structure is particularly useful when generating multiple responses per prompt:

\begin{Verbatim}
# Generate 3 independent responses to the same prompt
responses <- chat(
  prompts = "Generate 5 items measuring conscientiousness for a personality scale.",
  model = "OpenAI/gpt-5.4",
  openai.API = "your-key",
  reps = 3
)

# Each row is an independent generation
nrow(responses)  # 3
\end{Verbatim}

You can also customize the model's behavior using a \emph{system role.} When interacting with an LLM, there are typically two types of input: 1) the \textbf{user prompt} (the primary instructions for the model), and 2) the \textbf{system role} (a persona for the model to embody while generating text) \parencite{chen2025unleashing}. The system role acts as a sort of ``behavioral prior'' that situates the model's identity, and thus its expertise, tone, and decision-making before it ever reads the user's prompt.

For example, a system role like \emph{``You are an expert psychometrician and test developer specializing in personality assessment''} tells the model to approach item generation from the perspective of a domain expert. This primes the model toward domain-consistent language, professional tone, and psychometric conventions that a generic prompt alone might not elicit.

If a system role is not provided, the package constructs one automatically based on the prompt-building parameters you supply (such as the scale title, target audience, and domain). However, we recommend that users draft their own system role. A good system role should be as specific as possible. For example:

\begin{Verbatim}
system.role <- "You are an expert scale developer and psychometrician
  with extensive expertise in drafting Likert-type items for children
  with ASD. Today, you will focus on developing robust, single-statement
  items that assess linguistic ability."
\end{Verbatim}

In addition to setting a custom system role, the user can add multiple prompts to the same request. Here is an example where multiple primary prompts are used in one function call:

\begin{Verbatim}
# Using a system role and multiple prompts
responses <- chat(
  prompts = c(
    "Generate 3 items about anxiety related to AI replacing jobs.",
    "Generate 3 items about anxiety related to AI making decisions."
  ),
  model = "sonnet",  # Anthropic Claude Sonnet
  anthropic.API = "your-key",
  system.role = "You are an expert psychometrician specializing in scale development.",
  max.tokens = 500
)

# The result has one row per prompt
nrow(responses)  # 2
\end{Verbatim}

In the above example, only one system role was specified, so the same system role was used for both prompts. However, two system roles could have been provided, in which case the first system role in the list would be sent alongside the first prompt and the second system role pairs with the second prompt.

\subsection{Generating Text Locally}\label{generating-text-locally}

The \texttt{local\_\allowbreak{}chat()} function provides the same interface but routes all inference through a locally installed GGUF model. No internet connection or API key is required:

\begin{Verbatim}
# Using a local model (requires prior download via get_local_llm)
response <- local_chat(
  prompts = "Generate 5 items measuring conscientiousness for a personality scale.",
  model.path = model_path,  # Path from get_local_llm()
  system.role = "You are an expert psychometrician.",
  n.gpu.layers = -1  # Use all available GPU layers
)

print(response)
\end{Verbatim}

The \texttt{local\_\allowbreak{}chat()} function supports the same parameters as \texttt{chat()}. The key additional parameter is \texttt{model.\allowbreak{}path}, which must point to a valid GGUF file.

\section{Part 3: Embeddings}\label{part3}

Text embeddings are central to the AI-GENIE methodology. Embeddings transform human language into a high-dimensional numeric vector that captures its semantic meaning, enabling mathematical operations and psychometric analyses on text itself. Within the AI-GENIE pipeline, these embeddings serve in place of item responses for all analyses (e.g., EGA network construction).

While the \texttt{AIGENIE()} and \texttt{GENIE()} functions handle embedding generation internally, it can be useful to generate and inspect embeddings independently. This section covers the available embedding providers, how to select among them, and when different options are most appropriate.

\subsection{Embedding Providers}\label{embedding-providers}

The package supports three providers (Table \ref{tab:embedding-providers}). As described in Section \ref{APIkeys}, these providers require an API key to access their models; API keys can be made by creating an account on the provider's website.

\begin{table}[ht]
\centering
\caption{Embedding model providers compatible with the AIGENIE package}
\label{tab:embedding-providers}
\begin{tabular}{p{0.15\textwidth} p{0.31\textwidth} p{0.23\textwidth} p{0.19\textwidth}}
\toprule
\textbf{Provider} & \textbf{Notes} & \textbf{Example} & \textbf{Website} \\
\midrule

\textbf{OpenAI} &
Offers state-of-the-art embedding models that have been extensively validated in the AI-GENIE simulations &
\texttt{text-embedding-3-small} (the package default) &
\url{https://platform.openai.com} \\

\textbf{HuggingFace} &
An extremely wide ecosystem of open-source embedding models. Most usage is free. &
\texttt{BAAI/\allowbreak{}bge-base-en-v1.5} &
\url{https://huggingface.co} \\

\textbf{Jina AI} &
Offers competitive open-source models. The first 10 million tokens are free. &
\texttt{jina-embeddings-v4} &
\url{https://jina.ai/embeddings} \\

\bottomrule
\end{tabular}
\end{table}

The choice of embedding model can affect the quality of the network constructed in subsequent steps. The original AI-GENIE simulations used OpenAI's \texttt{text-embedding-3-small}, which demonstrated strong performance across all conditions \parencite{russelllasalandra2026aigenie,garrido2026estimating}. For most users, this remains the recommended default.

\subsection{Generating Embeddings Independently}\label{generating-embeddings-independently}

Although the pipeline functions handle embeddings automatically, researchers may want to generate embeddings independently. For example, it may be useful to compare embedding models, to use embeddings in other analyses, or to embed items once and then run the GENIE pipeline multiple times with different settings. The \texttt{AIGENIE()} and \texttt{GENIE()} functions both provide an \texttt{embeddings.\allowbreak{}only} flag for this purpose:

\begin{Verbatim}
# Generate items and their embeddings, without running the reduction pipeline
result <- AIGENIE(
  item.attributes = list(
    neuroticism = c("anxious", "depressed", "insecure", "emotional")
  ),
  openai.API = "your-key",
  model = "OpenAI/gpt-5.4",
  embedding.model = "text-embedding-3-small",
  embeddings.only = TRUE
)

# extract the embeddings from the result object 
my_embedding_matrix <- result$embeddings

str(my_embedding_matrix) # the number of items = number of columns
\end{Verbatim}

\section{Part 4: Item Generation}\label{part4}

One of the core capabilities of the \texttt{AIGENIE} package is automated item generation. Given a set of psychological constructs and their attributes, the package prompts an LLM to generate novel candidate items.

\subsection{Two Prompting Modes: Built-In and Custom}\label{two-prompting-modes-built-in-and-custom}

\texttt{AIGENIE()} offers two distinct ``modes'' for how prompts are constructed and sent to the model when generating items using an LLM. The choice between them determines how much control the user has over the exact wording of the instructions the LLM receives.

Throughout this section, the Big Five personality example will be used again. Here is the \texttt{item.\allowbreak{}attributes} object that will be referenced in this section's examples:

\begin{Verbatim}
big5_attributes <- list(
    openness = c("creative", "perceptual", "curious", "philosophical"),
    conscientiousness = c("organized", "responsible", "disciplined", "prudent"),
    extraversion = c("friendly", "positive", "assertive", "energetic"),
    agreeableness = c("cooperative", "compassionate", "trustworthy", "humble"),
    neuroticism = c("anxious", "depressed", "insecure", "emotional")
    )
\end{Verbatim}

\subsubsection{Built-In Prompting Mode (Recommended)}\label{built-in-prompting-mode-recommended}

\begin{figure}[!htbp]
    \centering
    \includegraphics[width=0.95\linewidth]{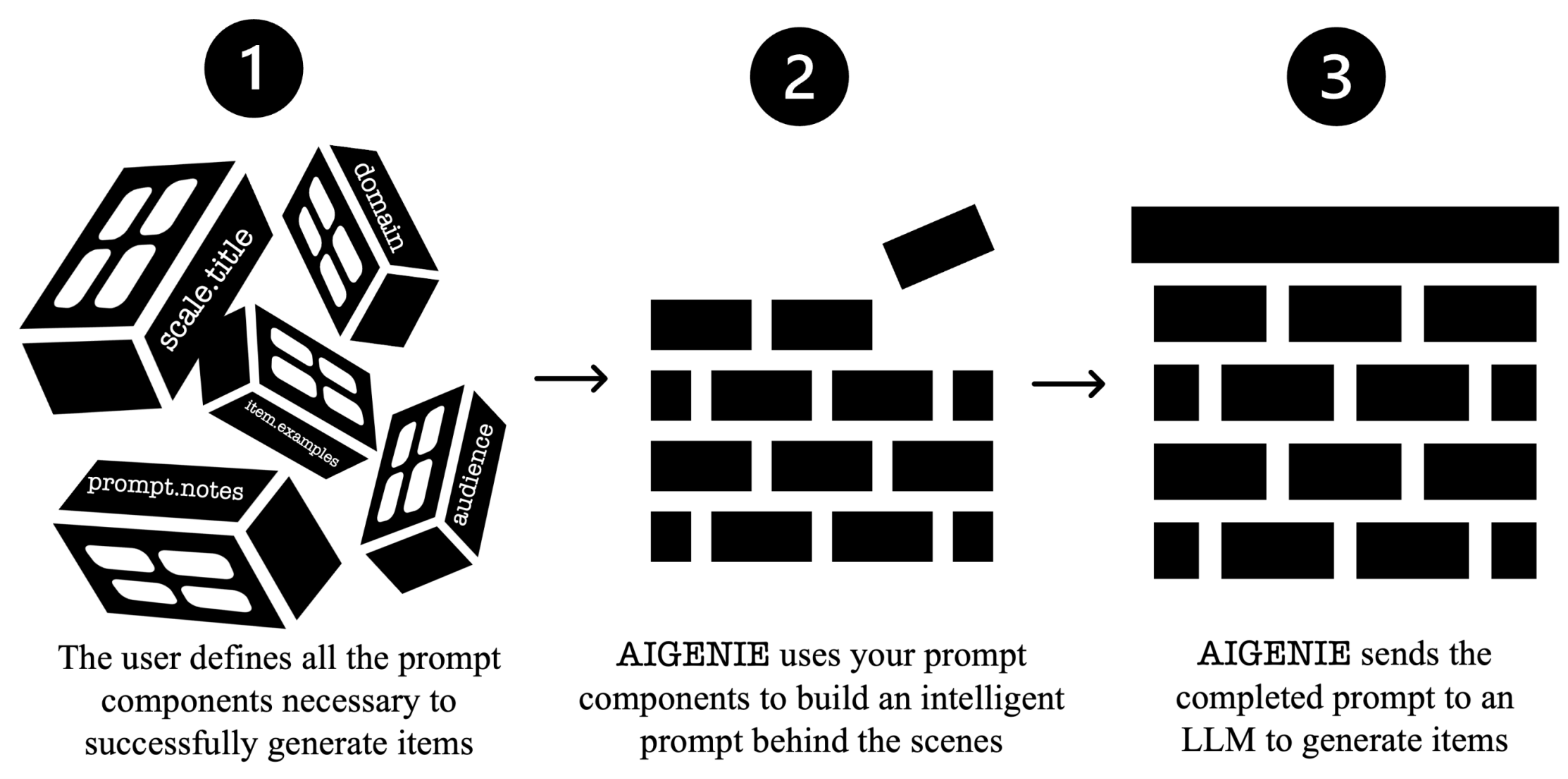}
    \caption{\textit{How \texttt{AIGENIE} Constructs a Prompt in Built-In Mode.} The user supplies descriptive components such as \texttt{domain}, \texttt{scale.\allowbreak{}title}, \texttt{audience}, \texttt{item.\allowbreak{}type.\allowbreak{}definitions}, \texttt{item.\allowbreak{}examples}, and \texttt{prompt.\allowbreak{}notes}. The package assembles these into a complete prompt and sends it to the LLM, so that the researcher specifies what the items should cover without writing the prompt itself.}
    \label{fig:fig-prompt-construction}
\end{figure}

In the \emph{built-in prompting mode} (the default), the user provides descriptive components and the package assembles them into a complete, well-structured prompt behind the scenes. These descriptive components are analogous to bricks; the \texttt{AIGENIE()} function arranges, aggregates, and assembles these ``bricks'' into a stable, coherent ``structure.'' This ``structure'' is the primary user prompt passed to the LLM. Figure \ref{fig:fig-prompt-construction} illustrates that assembly from component to finished prompt. This mode is recommended for most researchers because it incorporates the prompt engineering strategies shown to produce high-quality items in simulation studies \parencite{russelllasalandra2025prompt}.

The descriptive components (or ``bricks'') that the user can provide in the built-in prompting mode are all \emph{optional}, but users should provide as many as possible. These components are as follows:

\begin{itemize}
\tightlist
\item
  \texttt{domain} specifies the research domain (e.g., ``personality measurement'' or ``child development'').
\item
  \texttt{scale.\allowbreak{}title} provides the name of the scale being developed.
\item
  \texttt{audience} describes the scale's intended target population (e.g., ``college-educated adults'' or ``children with ASD in second grade'').
\item
  \texttt{item.\allowbreak{}type.\allowbreak{}definitions} is a named list providing a brief definition of each item type, giving the model substantive context about the construct.
\item
  \texttt{item.\allowbreak{}attribute.\allowbreak{}definitions} is a named list providing a brief definition of at least one of the item attributes. While we anticipate that attributes will be plain-language descriptors of informal aspects of the item type most of the time, there may be cases where complex attributes warrant further explication or precision.
\item
  \texttt{response.\allowbreak{}options} specifies the intended response options for the items.
\item
  \texttt{item.\allowbreak{}examples} is a data frame containing high-quality example items to guide the model's generation style and format. This data frame should contain \texttt{statement}, \texttt{type}, and \texttt{attribute} columns. The \texttt{statement} column should contain the item's actual content and the \texttt{type}/\texttt{attribute} columns are self-explanatory.
\item
  \texttt{prompt.\allowbreak{}notes} allows the user to append custom instructions to the automatically constructed prompt without having to write the entire prompt from scratch (e.g., ``ensure every item begins with the stem `I am someone who\ldots{}'\,'' or ``items should be very brief and contain no words that would exceed a fifth-grade vocabulary''). That is, the user can inject specific requirements or constraints while still benefiting from the package's built-in prompt engineering. If the built-in prompt handles most of what you need and only a small adjustment is required, \texttt{prompt.\allowbreak{}notes} is the right tool for the job.
\end{itemize}

Here is what an \texttt{AIGENIE()} function call might look like if using the built-in prompting mode:

\begin{Verbatim}
# Generate items using the built-in prompt
items_builtin <- AIGENIE(
  
  item.attributes = big5_attributes, # defined in previous examples
  openai.API = "sk-your-API-KEY", 
  model = "OpenAI/gpt-5.4", 

  # Descriptive components for prompt construction (BRICKS)
  domain = "personality measurement",
  scale.title = "Big Five Personality Inventory",
  audience = "college-educated adults in the United States",
  item.type.definitions = list(
      openness = "Openness reflects intellectual curiosity, aesthetic sensitivity, 
      and a preference for novelty and variety.",
      conscientiousness = "Conscientiousness reflects a tendency toward
      self-discipline, goal-directed behavior, and organization.",
      extraversion = "Extraversion reflects sociability, assertiveness, 
      and the tendency to seek stimulation in the company of others.",
      agreeableness = "Agreeableness reflects a tendency to be cooperative, 
      compassionate, and trusting toward others.",
      neuroticism = "Neuroticism reflects emotional instability, including proneness 
      to anxiety, sadness, and mood swings."
    ),
  response.options = c("strongly disagree", "disagree",
                      "neutral", "agree", "strongly agree"),
  prompt.notes = "All items should be written as first-person
      self-report statements beginning with 'I am someone who'.",
  
  # System role... the model's persona
  system.role = "You are an expert psychometrician and test
      developer specializing in personality assessment.",
  
  target.N = 8, # Generating only 8 items per item type
  items.only = TRUE # ONLY generating items in this example
)
\end{Verbatim}

\subsubsection{Custom Prompting Mode}\label{custom-prompting-mode}

In the \emph{custom prompting mode}, the user supplies fully written prompts via the \texttt{main.\allowbreak{}prompts} parameter. In this mode, the user has complete control over the exact instructions the model receives.

However, custom prompting is substantially more demanding than the built-in mode. When the package constructs prompts automatically, it includes several components that are easy to overlook when writing prompts from scratch. Users drafting custom prompts must include clear task framing, explicit mention of all attributes, proper incorporation of example items, and much more. Omitting any of these components can potentially lead to LLM outputs that are unparseable. For this reason, we recommend that most researchers start with the built-in mode and only switch to custom prompting after gaining familiarity with the package and with prompt engineering more broadly.

For researchers who do choose custom prompting, there are several requirements:

\begin{enumerate}
\def\labelenumi{\arabic{enumi}.}
\tightlist
\item
  The \texttt{main.\allowbreak{}prompts} parameter must be a \emph{named list with one prompt per item type}, and the names must match the names in the \texttt{item.\allowbreak{}attributes} parameter exactly.
\item
  Each prompt \emph{must} explicitly reference all of the attributes listed in the corresponding element of \texttt{item.\allowbreak{}attributes}.
\item
  Each prompt should be self-contained. That is, it should provide \emph{all} the context the model needs for that particular item type, because each prompt is sent to the model independently.
\end{enumerate}

A well-constructed custom prompt typically includes the following components, in roughly this order:

\begin{itemize}
\tightlist
\item
  Specific task context (i.e., what is being generated and why).
\item
  A definition of the target construct.
\item
  Explicit instructions about how many items to generate and how they should be distributed across attributes.
\item
  A list of the attributes, named exactly as they appear in \texttt{item.\allowbreak{}attributes}.
\item
  Quality constraints (e.g., instructions to generate novel items or avoid existing measures).
\end{itemize}

For example, custom prompts for the Big Five personality example may look like this:

\begin{Verbatim}
custom.main.prompts <- list(
  openness = "You are generating novel items targeting the Big Five
    personality trait openness to experience. Openness to experience
    is a personality trait that describes how open-minded, creative,
    and imaginative a person is. Generate EXACTLY eight items total
    for openness to experience; generate two items per attribute of
    openness to experience. These attributes are as follows: 1)
    creative, 2) perceptual, 3) curious, and 4) philosophical. Do
    NOT add or remove any attributes; use the attributes EXACTLY as
    provided. All items should be first-person self-report
    statements beginning with 'I am someone who'. Do NOT look for
    items that already exist in the literature; all items should
    be novel. Don't be afraid to push the bounds of the construct.",

  conscientiousness = "You are generating novel items targeting the
    Big Five personality trait conscientiousness. Conscientiousness
    is a personality trait that describes one's tendency toward self-discipline, 
    goal-directed behavior, and organization. Generate
    EXACTLY eight items total for conscientiousness; generate two
    items per attribute of conscientiousness. These attributes are as
    follows: 1) organized, 2) responsible, 3) disciplined, and 4)
    prudent. Do NOT add or remove any attributes; use the attributes
    EXACTLY as provided. All items should be first-person self-report
    statements beginning with 'I am someone who'. Do NOT look
    for items that already exist in the literature; all items should
    be novel. Don't be afraid to push the bounds of the construct.",
  
  extraversion = "You are generating novel items targeting the Big
    Five personality trait extraversion. Extraversion is a personality
    trait that describes people who are more focused on the external
    world than their internal experience. Generate EXACTLY eight
    items total for extraversion; generate two items per attribute of
    extraversion. These attributes are as follows: 1) friendly, 2)
    positive, 3) assertive, and 4) energetic. Do NOT add or remove
    any attributes; use the attributes EXACTLY as provided. All items
    should be first-person self-report statements beginning with 'I
    am someone who'. Do NOT look for items that already exist in the
    literature; all items should be novel. Don't be afraid to push
    the bounds of the construct.",

agreeableness = "You are generating novel items targeting the Big
    Five personality trait agreeableness. Agreeableness is a
    personality trait that describes one's tendency to be
    cooperative, compassionate, and trusting toward others. Generate
    EXACTLY eight items total for agreeableness; generate two items
    per attribute of agreeableness. These attributes are as follows:
    1) cooperative, 2) compassionate, 3) trustworthy, and 4) humble.
    Do NOT add or remove any attributes; use the attributes EXACTLY
    as provided. All items should be first-person self-report
    statements beginning with 'I am someone who'. Do NOT look for
    items that already exist in the literature; all items should be
    novel. Don't be afraid to push the bounds of the construct.",
  
  neuroticism = "You are generating novel items targeting the Big Five
    personality trait neuroticism. Neuroticism is a personality trait
    that describes one's tendency to experience negative emotions
    like anxiety, depression, irritability, anger, and self-consciousness. 
    Generate EXACTLY eight items total for neuroticism; generate two items per 
    attribute of neuroticism. These attributes are as follows: 1) anxious, 
    2) depressed, 3) insecure, and 4) emotional. Do NOT add or remove any 
    attributes; use the attributes EXACTLY as provided. All items should be first-person
    self-report statements beginning with 'I am someone who'.
    Do NOT look for items that already exist in the literature; all
    items should be novel. Don't be afraid to push the bounds of the construct."
)
\end{Verbatim}

Notice that each prompt follows the same general structure but is tailored to the specific construct. The attributes are listed by name exactly as they appear in \texttt{item.\allowbreak{}attributes}, the number of items per attribute is specified explicitly, and quality constraints are included. If even one attribute is misspelled or omitted, the package will not be able to parse and label the resulting items correctly. Moreover, if the researcher later decides to change the item stem from ``I am someone who'' to, say, ``I tend to,'' that change must be applied manually across all five prompts in the custom mode. In the built-in mode, it requires editing only the single \texttt{prompt.\allowbreak{}notes} string.

Here is how this \texttt{main.\allowbreak{}prompts} object might be used in an \texttt{AIGENIE()} function call:

\begin{Verbatim}
items_custom <- AIGENIE(
  main.prompts = custom.main.prompts, 
  item.attributes = big5_attributes,  
  openai.API = "sk-your-API-KEY", 
  model = "OpenAI/gpt-5.4", 
  
  # System role... the model's persona
  system.role = "You are an expert psychometrician and test
      developer specializing in personality assessment.",
  
  target.N = 8, # Generating only 8 items per item type
  items.only = TRUE # ONLY generating items in this example
  
)
\end{Verbatim}

Notice that this call does not include any ``brick'' parameters like \texttt{audience}, \texttt{scale.\allowbreak{}title}, \texttt{domain}, or \texttt{prompt.\allowbreak{}notes} like in the previous built-in prompt example. The ``bricks'' are unnecessary if you already have a sturdy ``wall/structure'' made. If these ``brick'' parameters are supplied in the custom prompting mode, they will be ignored by the function.

\section{Part 5: The Full AI-GENIE Pipeline}\label{part5}

The full AI-GENIE pipeline automates the entire workflow from item generation through structural validation. A single call to the \texttt{AIGENIE()} function generates items, embeds them, estimates the network structure, removes redundant items, assesses stability, and returns a validated item pool. This section walks through the complete pipeline using two examples: a simple Big Five demonstration and a more extended AI Anxiety application.

\subsection{Example 1: Big Five Personality}\label{example-1-big-five-personality}

We begin with a focused example generating items for three of the Big Five traits (in the interest of brevity, not all five personality traits were used in this example):

\begin{Verbatim}
# Define attributes for three personality traits
personality_attributes <- list(
  openness = c("creative", "perceptual", "curious", "philosophical"),
  conscientiousness = c("organized", "responsible", "disciplined", "prudent"),
  neuroticism = c("anxious", "depressed", "insecure", "emotional")
)

# Run the full AI-GENIE pipeline
abv_results <- AIGENIE(
  item.attributes = personality_attributes,
  openai.API = "your-openai-key",
  model = "OpenAI/gpt-5.4",
  embedding.model = "text-embedding-3-small",
  EGA.model = NULL, # test both TMFG and glasso; keep the higher NMI
  domain = "personality measurement",
  scale.title = "Big Five Personality Inventory",
  item.type.definitions = list(
    openness = "Openness reflects intellectual curiosity, aesthetic sensitivity, 
    and preference for novelty.",
    conscientiousness = "Conscientiousness reflects self-discipline, 
    goal-directed behavior, and organization.",
    neuroticism = "Neuroticism reflects emotional instability and proneness to 
    anxiety and mood swings."
  ),
  target.N = 60
)
\end{Verbatim}

\subsection{Understanding the Output}\label{understanding-the-output}

This section describes the default output that most researchers will see when they work with \texttt{AIGENIE} (i.e., the object returned when \texttt{items.\allowbreak{}only}, \texttt{embeddings.\allowbreak{}only}, \texttt{run.\allowbreak{}overall}, \texttt{keep.\allowbreak{}org}, and \texttt{all.\allowbreak{}together} are all left at their default value of \texttt{FALSE}). Variations introduced by changing each flag from its default are described at the end of this section.

\subsubsection{Top-Level Structure}\label{top-level-structure}

The default output is a named list with two top-level elements, \texttt{item\_\allowbreak{}type\_\allowbreak{}level} and \texttt{filtering\_\allowbreak{}audit}:

\begin{Verbatim}
names(abv_results)
#> [1] "item_type_level" "filtering_audit"
\end{Verbatim}

The \texttt{item\_\allowbreak{}type\_\allowbreak{}level} object is itself a named list containing one element per item type, named to match \texttt{item.\allowbreak{}attributes}. Recall that, by default, only items of like type run through the pipeline together. Thus, there are as many independent pipeline reduction results as there are item types.

In our abbreviated example where we generated items for \emph{three} personality traits, the \texttt{abv\_\allowbreak{}results\$item\_\allowbreak{}type\_\allowbreak{}level} return object contains \emph{three} sublists. These sublists correspond to each of the \emph{three} traits: \texttt{openness}, \texttt{conscientiousness}, and \texttt{neuroticism}. Each sublist holds the complete set of results from the reduction pipeline as applied to that item type in isolation.

\begin{Verbatim}
names(abv_results$item_type_level)
#> [1] "openness"          "conscientiousness" "neuroticism"
\end{Verbatim}

\subsubsection{Results on the Item Type Level}\label{results-on-the-item-type-level}

Since items are evaluated in the reduction pipeline completely agnostic of items that are of other types, separate and independent results are available for each item type. In other words, conscientiousness items go through the reduction pipeline only with other conscientiousness items, neuroticism items go through the pipeline only with other neuroticism items, and so on. Each per-type sublist within \texttt{abv\_\allowbreak{}results\$item\_\allowbreak{}type\_\allowbreak{}level} contains fifteen elements. In this example, then, there would be fifteen result elements pertaining to each of the three personality traits. These are the fifteen elements:

\begin{Verbatim}
# Pick a trait (openness) to see what results are available on the
# item-type level in general
names(abv_results$item_type_level$openness)
#>  [1] "final_NMI"          "initial_NMI"        "embeddings"
#>  [4] "UVA"                "bootEGA"            "EGA.model_selected"
#>  [7] "final_items"        "final_EGA"          "initial_EGA"
#> [10] "start_N"            "final_N"            "network_plot"
#> [13] "stability_plot"     "filtering_audit"    "reduction_summary"
\end{Verbatim}

Here is a closer look at each of these fifteen elements:

\begin{itemize}
\item
  \texttt{final\_\allowbreak{}items} is the most practically important element. It is a \texttt{data.\allowbreak{}frame} containing the items that survived the full reduction pipeline. Its columns are:

  \begin{itemize}
  \tightlist
  \item
    \texttt{ID}: A numeric identifier assigned during generation.
  \item
    \texttt{statement}: The item's actual text (e.g., ``I enjoy combining unrelated ideas into something original'').
  \item
    \texttt{attribute}: The attribute the item was generated to reflect (e.g., \texttt{"creative"}).
  \item
    \texttt{type}: The item type the item belongs to (e.g., \texttt{"openness"}).
  \item
    \texttt{EGA\_\allowbreak{}com}: The community assignment from the final EGA network. Comparing \texttt{attribute} against \texttt{EGA\_\allowbreak{}com} is the most useful inspection at this stage.
  \end{itemize}
\item
  \texttt{start\_\allowbreak{}N} records the total number of items generated for this item type. In our example, \texttt{abv\_\allowbreak{}results\$item\_\allowbreak{}type\_\allowbreak{}level\$openness\$start\_\allowbreak{}N} returns \texttt{64}. Thus, there were 64 openness items in the initial item pool before reduction. Note that \texttt{target.\allowbreak{}N} was set to 60; generation proceeds in batches and stops once the target is met, so the final count is approximate and often slightly exceeds it.
\item
  \emph{\texttt{final\_\allowbreak{}N}} records the total number of items remaining for this item type after reduction. In our example, \texttt{abv\_\allowbreak{}results\$item\_\allowbreak{}type\_\allowbreak{}level\$openness\$final\_\allowbreak{}N} returns \texttt{61}.
\item
  \texttt{initial\_\allowbreak{}NMI} is a numeric value reporting the NMI between the EGA network-detected community structure and the known attribute assignments before any reduction steps. For openness, this was \texttt{0.915}.
\item
  \texttt{final\_\allowbreak{}NMI} is a numeric value reporting the NMI between the EGA network-detected community structure and the known attribute assignments after all reduction steps. For openness, this rose to \texttt{0.956}.
\item
  \texttt{reduction\_\allowbreak{}summary} is a \texttt{data.\allowbreak{}frame} with one row per stage of the pipeline, summarizing the trajectory of the reduction in a single table. Its columns are \texttt{stage}, \texttt{N} (items remaining), \texttt{NMI}, \texttt{n\_\allowbreak{}removed\_\allowbreak{}at\_\allowbreak{}stage}, and \texttt{delta\_\allowbreak{}NMI} (the change in NMI relative to the initial model).
\item
  \texttt{UVA} is a list containing results from the UVA redundancy reduction step. UVA identifies pairs or sets of items whose embeddings are so similar that retaining them all would introduce redundancy. An item is flagged when its weighted topological overlap with another item reaches the \texttt{uva.\allowbreak{}cut.\allowbreak{}off}, which defaults to \texttt{0.20}. The list contains:

  \begin{itemize}
  \tightlist
  \item
    \texttt{n\_\allowbreak{}removed} is the total number of redundant items removed across all UVA sweeps.
  \item
    \texttt{n\_\allowbreak{}sweeps} is the number of iterative passes UVA required before no further redundancies were detected.
  \item
    \texttt{redundant\_\allowbreak{}pairs} is a \texttt{data.\allowbreak{}frame} logging every redundancy decision. Each row records the sweep in which the redundancy was detected, the items involved, which item was kept, and which item or items were removed.
  \end{itemize}
\item
  \texttt{bootEGA} is a list containing results from the bootstrap EGA step. After UVA removes semantically redundant items, bootEGA evaluates the structural stability of the remaining items by repeatedly resampling the embedding matrix and estimating the network structure. The number of resamples is controlled by \texttt{boot.\allowbreak{}iter}, which defaults to \texttt{500}. Items that frequently change community membership across bootstrap samples are pruned. The list contains:

  \begin{itemize}
  \tightlist
  \item
    \texttt{post\_\allowbreak{}uva\_\allowbreak{}initial\_\allowbreak{}boot} is the \texttt{bootEGA} object (from the \texttt{EGAnet} package) estimated on the post-UVA item pool, before any stability-based pruning.
  \item
    \texttt{post\_\allowbreak{}uva\_\allowbreak{}final\_\allowbreak{}boot} is the \texttt{bootEGA} object estimated after stability-based pruning.
  \item
    \texttt{n\_\allowbreak{}removed} reports the number of items removed due to low structural stability across all bootEGA stability sweeps.
  \item
    \texttt{items\_\allowbreak{}removed} is a \texttt{data.\allowbreak{}frame} logging the specific items identified and pruned during the stability check.
  \item
    \texttt{initial\_\allowbreak{}boot\_\allowbreak{}with\_\allowbreak{}redundancies} is a \texttt{bootEGA} object estimated on the full, pre-UVA item pool. This object is useful for comparing the stability of the item pool before and after redundancy reduction.
  \end{itemize}
\item
  \texttt{filtering\_\allowbreak{}audit} is a \texttt{data.\allowbreak{}frame} giving a record of every removal for this item type, and is the single most useful object for explaining a reduction to a reader or reviewer. Each row is one removed item, and its twenty-five columns answer four questions in turn: what was removed (\texttt{ID}, \texttt{type}, \texttt{attribute}, \texttt{statement}); why it was removed (\texttt{removal\_\allowbreak{}stage}, \texttt{reason}, \texttt{diagnostic\_\allowbreak{}name}, \texttt{diagnostic\_\allowbreak{}value}, \texttt{cutoff}); what evidence was involved, whether for redundancy (\texttt{uva\_\allowbreak{}sweep}, \texttt{redundant\_\allowbreak{}with\_\allowbreak{}ID}, \texttt{redundant\_\allowbreak{}with\_\allowbreak{}statement}, \texttt{redundant\_\allowbreak{}wTO}, \texttt{all\_\allowbreak{}redundant\_\allowbreak{}with\_\allowbreak{}IDs}, \texttt{all\_\allowbreak{}redundant\_\allowbreak{}wTO}) or for instability (\texttt{boot\_\allowbreak{}run}, \texttt{item\_\allowbreak{}stability}, \texttt{stability\_\allowbreak{}deficit}); and what the item's structural profile looked like beforehand (\texttt{pre\_\allowbreak{}reduction\_\allowbreak{}EGA\_\allowbreak{}community}, \texttt{pre\_\allowbreak{}reduction\_\allowbreak{}primary\_\allowbreak{}network\_\allowbreak{}loading}, \texttt{pre\_\allowbreak{}reduction\_\allowbreak{}strongest\_\allowbreak{}cross\_\allowbreak{}community}, \texttt{pre\_\allowbreak{}reduction\_\allowbreak{}strongest\_\allowbreak{}cross\_\allowbreak{}loading}, \texttt{pre\_\allowbreak{}reduction\_\allowbreak{}loading\_\allowbreak{}gap}, and their \texttt{\_\allowbreak{}abs} variants).

  The \texttt{reason} column is written to be read directly. For example, \texttt{Instability:\ item\ stability\ =\ 0.406\ \textless{}\ 0.75} means that a removal decision can be reported without reconstructing it from the underlying statistics. The pre-reduction columns are most useful for explaining a removal substantively; an item whose \texttt{pre\_\allowbreak{}reduction\_\allowbreak{}loading\_\allowbreak{}gap} is near zero was loading almost equally on two communities, which is usually why it proved unstable.
\item
  \texttt{EGA.\allowbreak{}model\_\allowbreak{}selected} is a character string indicating which network estimation model was selected (TMFG or glasso). When \texttt{EGA.\allowbreak{}model\ =\ NULL}, the package tests both models and selects whichever produces a higher NMI. If the user specifies a model via the \texttt{EGA.\allowbreak{}model} parameter, this field simply reflects that choice. Because each item type is analyzed independently, different types may select different models: in our example, openness and neuroticism selected glasso while conscientiousness selected TMFG.
\item
  \texttt{initial\_\allowbreak{}EGA} is an \texttt{EGA} object (from the \texttt{EGAnet} package) representing the network structure of the item pool before the reduction pipeline.
\item
  \texttt{final\_\allowbreak{}EGA} is an \texttt{EGA} object representing the network structure after the reduction pipeline.
\item
  \texttt{embeddings} is a list containing the embedding matrices used in the analysis. The list contains three elements:

  \begin{itemize}
  \tightlist
  \item
    \texttt{full} is the dense (full) embedding matrix for the final retained items. Rows are embedding dimensions and columns are items, with column names corresponding to item IDs.
  \item
    \texttt{sparse} is the sparsified embedding matrix for the final retained items. Sparsification zeroes out values in the middle 95\% of the distribution, retaining only the most extreme embedding entries.
  \item
    \texttt{selected} names the embedding type (either \texttt{full} or \texttt{sparse}) used in the analysis. The package tests both and selects whichever yields a higher NMI.
  \end{itemize}
\item
  \texttt{network\_\allowbreak{}plot} stores the \texttt{ggplot}/\texttt{patchwork} object displaying a side-by-side comparison of the EGA networks before and after reduction, with the initial network on the left and the final network on the right. An example of this network plot for \texttt{conscientiousness} is shown in Figure \ref{fig:fig-consc-network}.
\item
  \texttt{stability\_\allowbreak{}plot} is a \texttt{ggplot}/\texttt{patchwork} object comparing item stability before (the plot on the left) and after (the plot on the right) reduction. Item stability refers to how consistently each item is assigned to the same community across bootstrap samples. Higher stability values (closer to 1) indicate that an item reliably belongs to the dimension it is assigned to in the majority of bootstrap samples. An example of such a plot for \texttt{conscientiousness} is shown in Figure \ref{fig:fig-consc-stability}.
\end{itemize}

\begin{figure}[!htbp]
    \centering
    \includegraphics[width=0.95\linewidth]{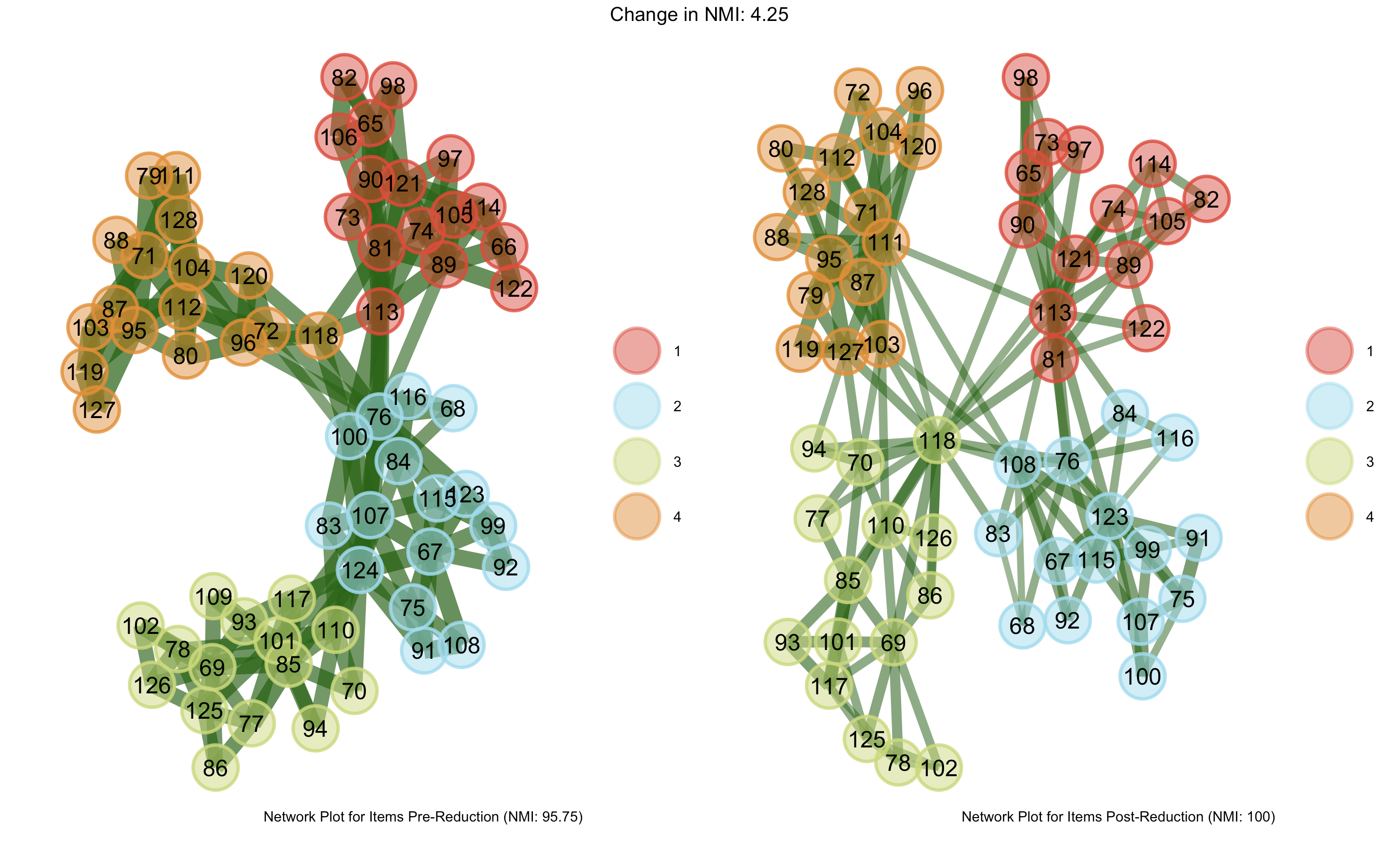}
    \caption{\textit{Conscientiousness Items: EGA Network Before and After Reduction.} The estimated network before item reduction (left) compared with the network after reduction (right). Node color indicates the community assigned by EGA.}
    \label{fig:fig-consc-network}
\end{figure}

\begin{figure}[!htbp]
    \centering
    \includegraphics[width=0.95\linewidth]{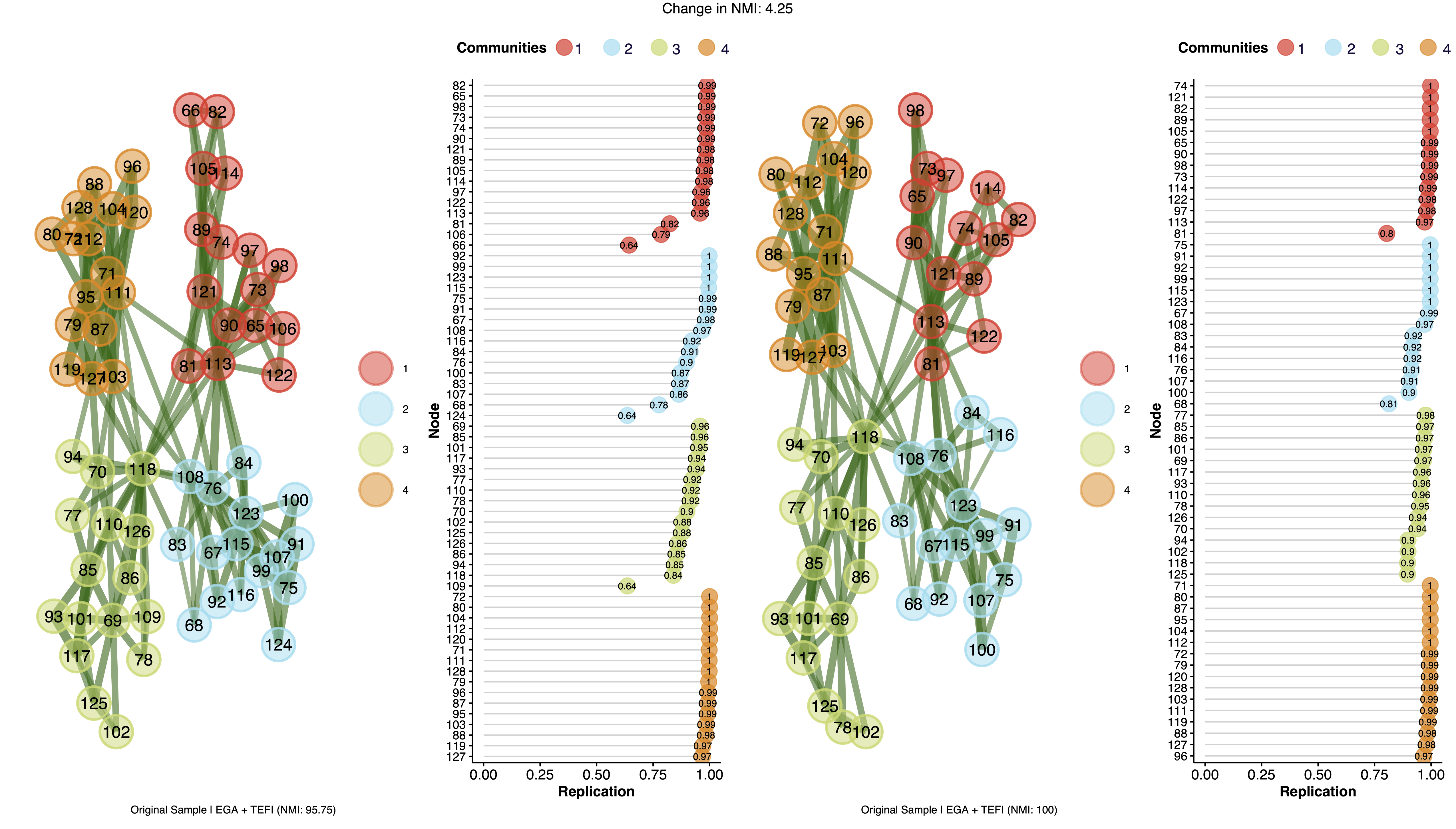}
    \caption{\textit{Conscientiousness Items: Item Stability Before and After Reduction.} Item stability before reduction (left) versus after reduction (right). Stability refers to how consistently each item is assigned to the same community across bootstrap samples, with values closer to 1 indicating more reliable assignment. Stability improves after reduction, and NMI increases by roughly 4 percentage points.}
    \label{fig:fig-consc-stability}
\end{figure}

\subsubsection{Output Variations by Flag}\label{output-variations-by-flag}

The aforementioned output structure occurs under default settings and is likely sufficient for many researchers.

However, this output structure may change, depending on whether any of these four flags are changed from the default of \texttt{FALSE} to \texttt{TRUE}: \texttt{items.\allowbreak{}only}, \texttt{embeddings.\allowbreak{}only}, \texttt{keep.\allowbreak{}org}, and \texttt{run.\allowbreak{}overall}. Additionally, the output changes when the flag \texttt{all.\allowbreak{}together\ =\ TRUE}, but in a more fundamental way; this flag is covered in Section \ref{alltogether}.

As discussed previously, when \texttt{items.\allowbreak{}only\ =\ TRUE}, the function skips embedding, network analysis, and reduction entirely. It returns a simple \texttt{data.\allowbreak{}frame} with four columns (\texttt{ID}, \texttt{statement}, \texttt{type}, and \texttt{attribute}). This \texttt{data.\allowbreak{}frame} contains the raw generated item pool.

When \texttt{embeddings.\allowbreak{}only\ =\ TRUE}, the function generates items and computes embeddings but skips the psychometric reduction pipeline. It returns a named list with two elements: \texttt{embeddings} (the embedding matrix) and \texttt{items} (the same items \texttt{data.\allowbreak{}frame} described above).

When \texttt{keep.\allowbreak{}org\ =\ TRUE}, the top-level structure remains the same (\texttt{item\_\allowbreak{}type\_\allowbreak{}level} and \texttt{filtering\_\allowbreak{}audit}), but each per-type sublist gains an \texttt{initial\_\allowbreak{}items} field. This field contains a data frame of all items generated \emph{before} reduction. Similarly, the embeddings sublists also gain \texttt{full\_\allowbreak{}org} and \texttt{sparse\_\allowbreak{}org} matrices corresponding to the full pre-reduction item pool. If \texttt{run.\allowbreak{}overall\ =\ TRUE} is also set, the \texttt{overall} element likewise gains an \texttt{initial\_\allowbreak{}items} data frame, and its own \texttt{embeddings} sublist gains \texttt{full\_\allowbreak{}org} and \texttt{sparse\_\allowbreak{}org}.

When \texttt{run.\allowbreak{}overall\ =\ TRUE}, the \texttt{item\_\allowbreak{}type\_\allowbreak{}level} results are unchanged, but the output gains an \texttt{overall} element containing a full analysis of the pooled item pool. It includes its own \texttt{final\_\allowbreak{}NMI}, \texttt{initial\_\allowbreak{}NMI}, \texttt{EGA.\allowbreak{}model\_\allowbreak{}selected}, \texttt{final\_\allowbreak{}EGA}, \texttt{initial\_\allowbreak{}EGA}, \texttt{start\_\allowbreak{}N}, \texttt{final\_\allowbreak{}N}, \texttt{filtering\_\allowbreak{}audit}, and \texttt{reduction\_\allowbreak{}summary}. It also includes a \emph{new} pooled network estimated across all item types at once (\texttt{network\_\allowbreak{}plot}), before and after reduction, which is displayed alongside the per-type plots when \texttt{plot\ =\ TRUE}.

It should be noted, however, that this overall analysis does \emph{not} perform additional reduction \emph{nor} does it change what items would have been removed. It only evaluates the combined post-reduction item pool as a whole in a post-hoc analysis.

\subsubsection{\texorpdfstring{The \texttt{all.\allowbreak{}together} Flag}{The all.together Flag}}\label{alltogether}

By default, the reduction pipeline runs separately for each item type, producing completely independent reduction results for each item type. Setting \texttt{all.\allowbreak{}together\ =\ TRUE} means all items run through the reduction pipeline together once.

Thus, the output structure when setting \texttt{all.\allowbreak{}together\ =\ TRUE} necessarily changes. Rather than the two-level \texttt{item\_\allowbreak{}type\_\allowbreak{}level} / \texttt{filtering\_\allowbreak{}audit} organization, the function returns a single flat named list containing: \texttt{final\_\allowbreak{}NMI}, \texttt{initial\_\allowbreak{}NMI}, \texttt{embeddings}, \texttt{UVA}, \texttt{bootEGA}, \texttt{EGA.\allowbreak{}model\_\allowbreak{}selected}, \texttt{final\_\allowbreak{}items}, \texttt{final\_\allowbreak{}EGA}, \texttt{initial\_\allowbreak{}EGA}, \texttt{start\_\allowbreak{}N}, \texttt{final\_\allowbreak{}N}, \texttt{network\_\allowbreak{}plot}, \texttt{stability\_\allowbreak{}plot}, \texttt{filtering\_\allowbreak{}audit}, and \texttt{reduction\_\allowbreak{}summary}. These are the same elements described in the per-type breakdown above, but applied to the combined pool.

\begin{Verbatim}
# Pool all Big Five items and reduce them as a single batch
abv_results_together <- AIGENIE(
  item.attributes = big5_attributes,
  openai.API = "your-key",
  model = "OpenAI/gpt-5.4",
  domain = "personality",
  scale.title = "Big Five Personality Inventory",
  all.together = TRUE
)

# The output is a flat list -- no item_type_level / filtering_audit split
names(abv_results_together)
#>  [1] "final_NMI"          "initial_NMI"        "embeddings"
#>  [4] "UVA"                "bootEGA"            "EGA.model_selected"
#>  [7] "final_items"        "final_EGA"          "initial_EGA"
#> [10] "start_N"            "final_N"            "network_plot"
#> [13] "stability_plot"     "filtering_audit"    "reduction_summary"

# The final_items data frame now contains items from ALL types
# with concatenated attributes (e.g., "openness creative")
abv_results_together$final_items

# View the single network encompassing all traits
abv_results_together$network_plot
\end{Verbatim}

It is important to note that \texttt{all.\allowbreak{}together} and \texttt{run.\allowbreak{}overall} are different parameters, though they have similar names. The \texttt{run.\allowbreak{}overall} flag does \emph{not} change how reduction works fundamentally. When \texttt{run.\allowbreak{}overall\ =\ TRUE}, the items are still pruned independently within each type by default. This flag simply adds a post-hoc EGA analysis of the combined final pool to assess cross-type structure. Crucially, this post-hoc analysis does \emph{not} change which items are pruned. With \texttt{all.\allowbreak{}together\ =\ TRUE}, however, the reduction itself operates on the complete, combined item pool, meaning that items can be removed because of cross-type redundancy or instability. Thus, \texttt{run.\allowbreak{}overall\ =\ TRUE} will answer ``what does the cross-type structure look like after per-type reduction?'', while \texttt{all.\allowbreak{}together\ =\ TRUE} will answer ``what happens when we let the reduction pipeline see all items at once?''

\subsection{Example Using AI Anxiety}\label{example-using-ai-anxiety}

We now turn to a much more realistic application of developing a scale for a nascent or emerging construct. Unlike the Big Five, AI Anxiety \parencite[AIA;][]{wang2022development} is \emph{not} a well-established construct with decades of psychometric research and extensive representation in LLM training data. AIA is an emerging construct that has only recently received formal measurement attention. \textcite{wang2022development} developed and validated the first dedicated AIA scale, identifying a four-factor structure based on distinct sources of anxiety that individuals experience in response to the development, deployment, and societal integration of artificial intelligence technologies. These four factors are:

\begin{enumerate}
\def\labelenumi{\arabic{enumi}.}
\item
  \emph{Learning anxiety} describes cognitive overwhelm in the face of AI complexity, including perceiving one's knowledge as insufficient for AI demands and feeling daunted by the pace of AI advancement.
\item
  \emph{Job replacement} describes fear of professional obsolescence, including anxiety that AI will eliminate one's professional role, belief that one's skills can be automated, and uncertainty about career stability.
\item
  \emph{Sociotechnical blindness} describes concern about societal AI impacts, including loss of autonomy to AI systems, concerns about AI-enabled privacy violations, and worry about over-reliance on AI technology.
\item
  \emph{AI configuration} describes anxiety about AI system opacity, including lacking confidence in AI decision-making, confusion about how AI systems operate, and doubting AI's reliability and safety.
\end{enumerate}

The original AI-GENIE paper included a small-scale simulation using this construct \parencite{russelllasalandra2026aigenie}, and we follow the same operationalization here. For each of the four factors, we define three attributes that capture the core themes of that factor:

\begin{Verbatim}
# Define the construct: AI Anxiety
# Following the four-factor structure from Wang & Wang (2022)
# as operationalized in Russell-Lasalandra et al. (2026)
ai_anxiety_attributes <- list(
  learning_anxiety = c("overwhelmed", "inadequacy", "intimidated"),
  job_replacement = c("threatened", "replaceable", "insecure"),
  sociotechnical_blindness = c("powerless", "overly dependent", "surveilled"),
  ai_configuration = c("distrustful", "uncertain", "vulnerable")
)

# Provide detailed definitions for each factor
ai_anxiety_definitions <- list(
  learning_anxiety = paste(
    "Learning anxiety refers to cognitive overwhelm in the face of AI complexity.",
    "It includes perceiving one's knowledge as insufficient for AI demands",
    "and feeling daunted by the pace of AI advancement."
  ),
  job_replacement = paste(
    "Job replacement anxiety refers to fear of professional obsolescence.",
    "It includes fearing that AI will eliminate one's professional role,",
    "believing one's skills can be automated, and experiencing uncertainty",
    "about career stability."
  ),
  sociotechnical_blindness = paste(
    "Sociotechnical blindness refers to concern about societal AI impacts.",
    "It includes loss of autonomy to AI systems, concerns about AI-enabled",
    "privacy violations, and worrying about over-reliance on AI technology."
  ),
  ai_configuration = paste(
    "AI configuration anxiety refers to anxiety about AI system opacity.",
    "It includes lacking confidence in AI decision-making, confusion about",
    "how AI systems operate, and doubting AI's reliability and safety."
  )
)

# Run the full pipeline
ai_anxiety_results <- AIGENIE(
  item.attributes = ai_anxiety_attributes,
  openai.API = "your-openai-key",
  model = "OpenAI/gpt-5.1",
  embedding.model = "text-embedding-3-small",
  EGA.model = NULL, # test both TMFG and glasso; keep the higher NMI
  domain = "technology-related psychological assessment",
  scale.title = "AI Anxiety Scale",
  audience = "adults who use or are exposed to AI technologies in daily life",
  item.type.definitions = ai_anxiety_definitions,
  response.options = c("strongly disagree", "disagree", "slightly disagree",
                       "slightly agree", "agree", "strongly agree"),
  target.N = 80,
  adaptive = TRUE,
  keep.org = TRUE,  # Keep the original (pre-reduction) items for inspection
  plot = TRUE
)
\end{Verbatim}

\begin{figure}[!htbp]
    \centering
    \includegraphics[width=0.95\linewidth]{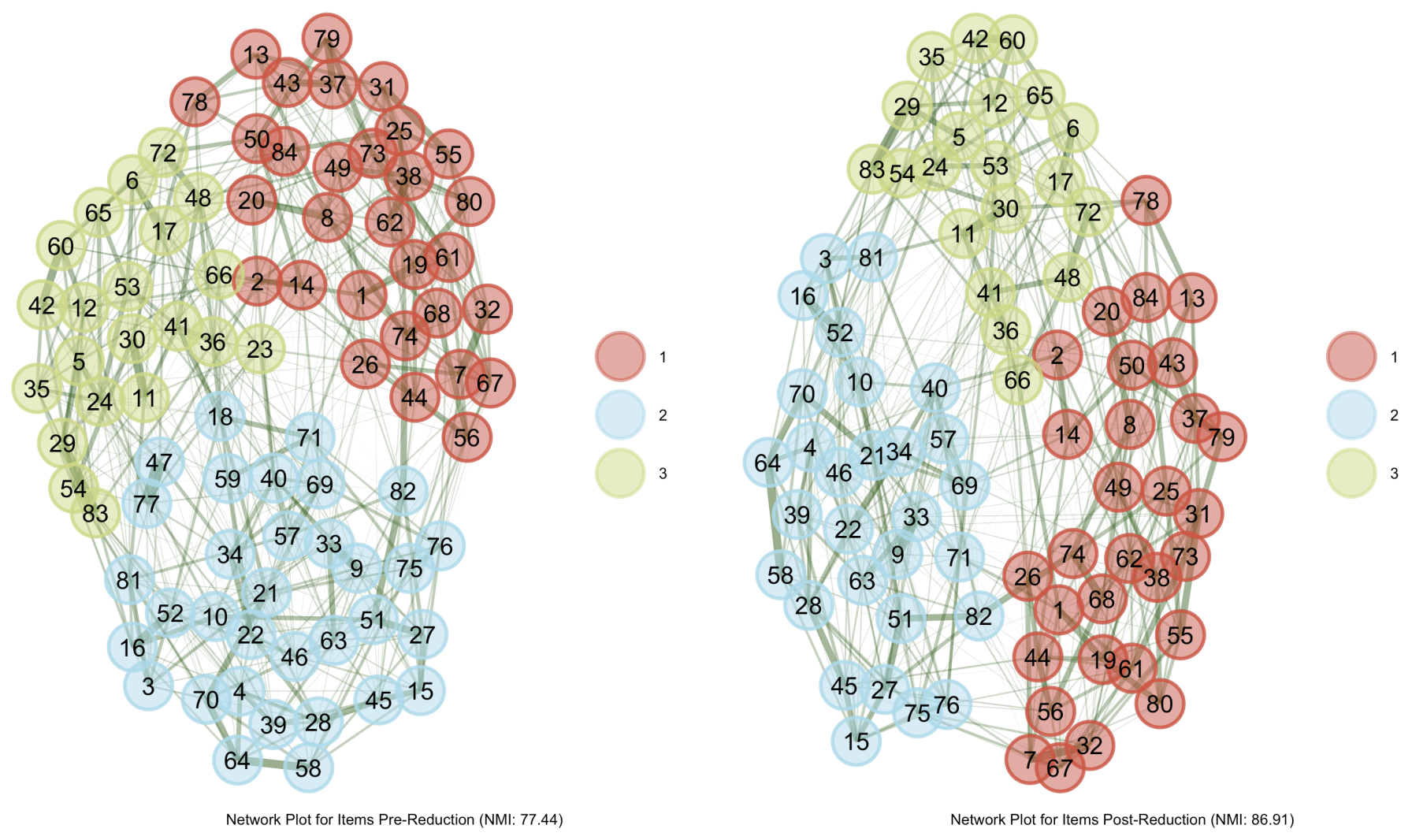}
    \caption{\textit{Learning Anxiety Items: EGA Network Before and After Reduction.} The estimated network before reduction (left) versus after reduction (right). NMI improved by 9.48 percentage points, reaching a final value of 86.91\%.}
    \label{fig:fig-learning-network}
\end{figure}

\begin{figure}[!htbp]
    \centering
    \includegraphics[width=0.95\linewidth]{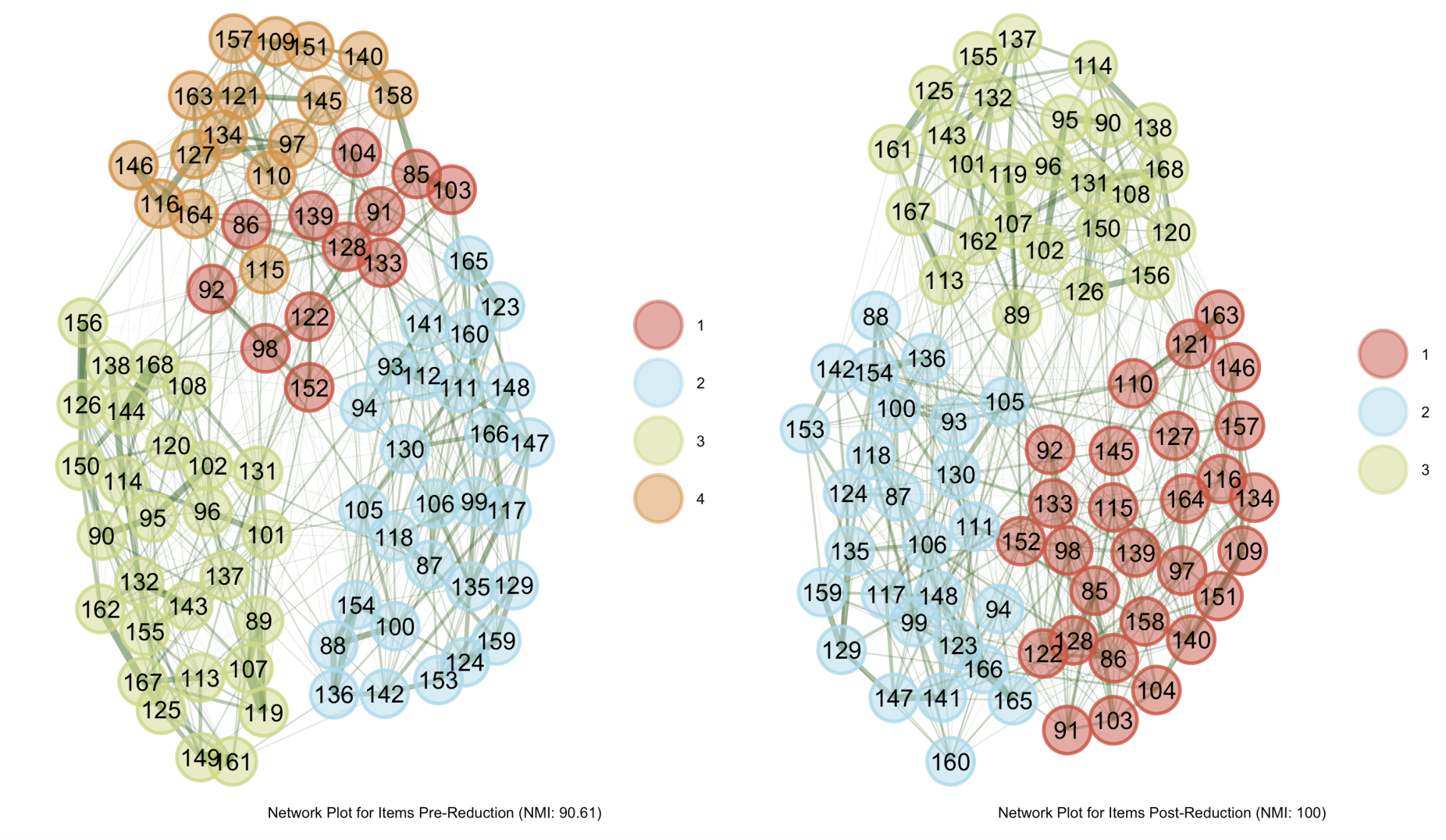}
    \caption{\textit{Job Replacement Items: EGA Network Before and After Reduction.} The estimated network before reduction (left) versus after reduction (right). NMI improved by 9.39 percentage points, reaching a final value of 100\%.}
    \label{fig:fig-jobrep-network}
\end{figure}

\begin{figure}[!htbp]
    \centering
    \includegraphics[width=0.95\linewidth]{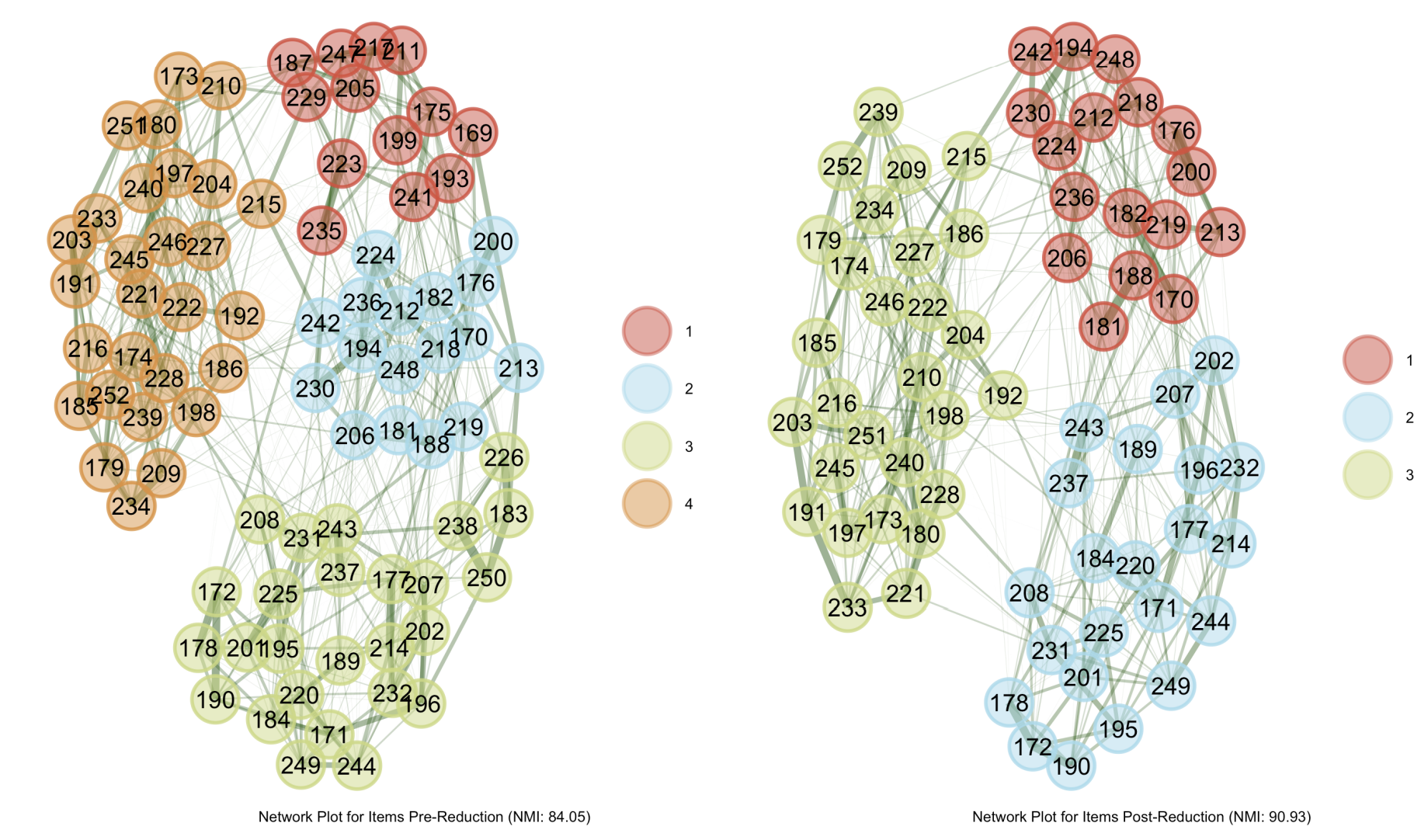}
    \caption{\textit{Sociotechnical Blindness Items: EGA Network Before and After Reduction.} The estimated network before reduction (left) versus after reduction (right). NMI improved by 6.88 percentage points, reaching a final value of 90.93\%.}
    \label{fig:fig-sociotech-network}
\end{figure}

\begin{figure}[!htbp]
    \centering
    \includegraphics[width=0.95\linewidth]{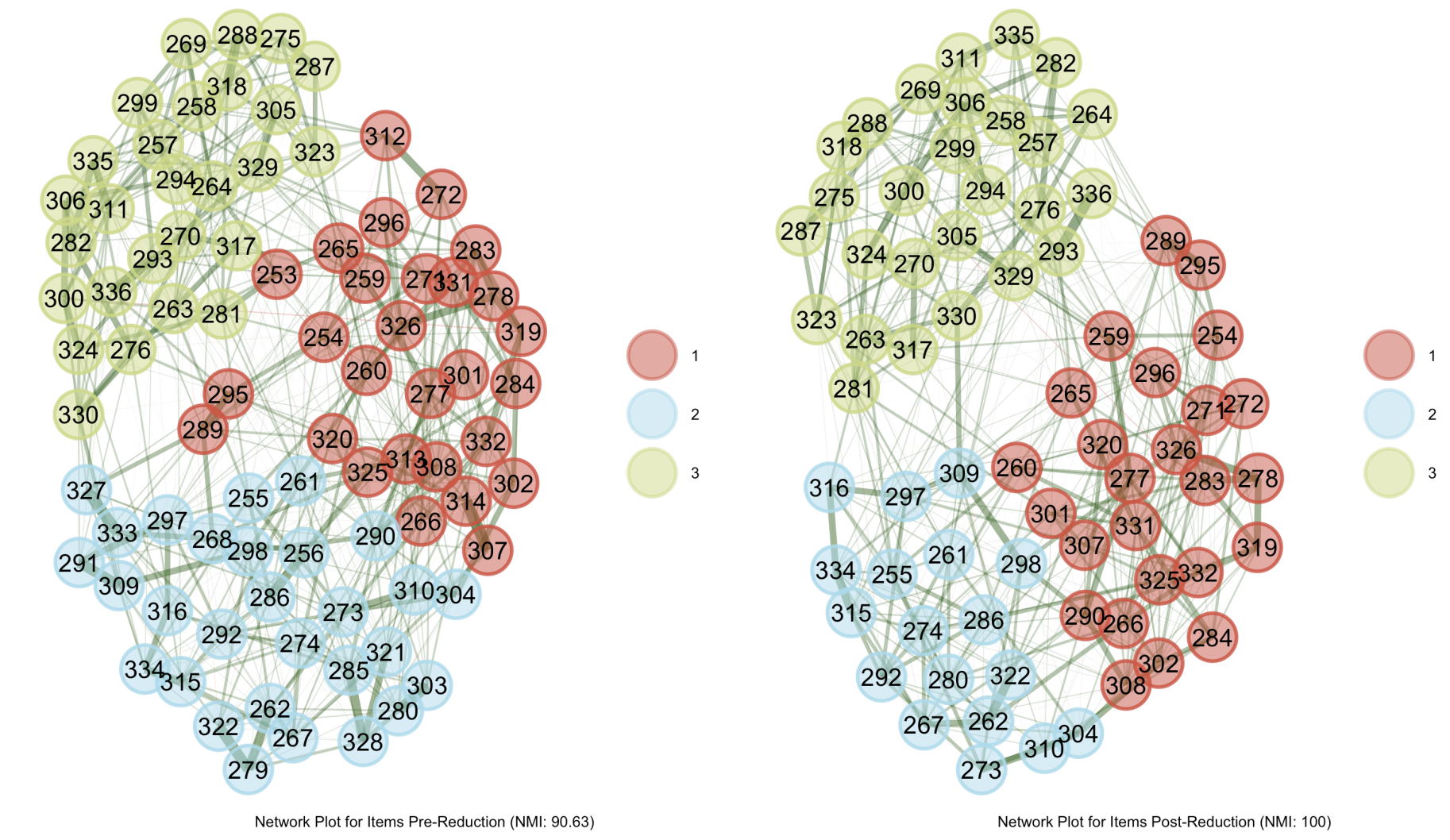}
    \caption{\textit{AI Configuration Items: EGA Network Before and After Reduction.} The estimated network before reduction (left) versus after reduction (right). NMI improved by 9.37 percentage points, reaching a final value of 100\%.}
    \label{fig:fig-aiconfig-network}
\end{figure}

The results from this single run were promising, as NMI trended upward after reduction for every factor. The network plots for each of the four factors are shown in Figures \ref{fig:fig-learning-network} through \ref{fig:fig-aiconfig-network}, each comparing the EGA network before reduction (left) with the network after reduction (right).

\section{Part 6: The GENIE Pipeline --- Psychometric Evaluation of Existing Items}\label{part6}

Not all researchers need AI-generated items, hence, the advent of \texttt{GENIE()}. This function performs all the same reduction steps as \texttt{AIGENIE()} without the LLM-generation step (that is, AIGENIE without ``AI''). The \texttt{GENIE()} function (or its local equivalent \texttt{local\_\allowbreak{}GENIE()}) applies the full network psychometric evaluation pipeline to any user-supplied set of items, without generating any new content.

The reduction pipeline itself is \emph{deterministic} and \emph{source-agnostic}. It is \emph{deterministic} in that the reduction itself introduces no randomness beyond the bootstrap resampling in bootEGA. Thus, two \texttt{GENIE()} runs on the same item pool, the same embedding matrix, and the same seed (done implicitly through the package) yield the \emph{same} retained items every time. Consequently, a published reduction can be reproduced exactly by anyone holding those objects. \texttt{GENIE()} is \emph{source-agnostic} in that the pipeline never sees \emph{how} an item came to exist. It operates only on the items and/or embedding matrix, and so evaluates items written by an LLM, drafted by an expert panel, or adapted from an existing instrument.

\subsection{A Reproducible Worked Example}\label{a-reproducible-worked-example}

The package ships an item pool and its matching embeddings so that the full reduction pipeline can be run without an API key or any generation step. This is the fastest way to see what \texttt{GENIE()} does before committing your own items to it. This built-in item pool example contains 180 LLM-generated personality items. More specifically, this pool contains 90 conscientiousness items spanning self-efficacy, achievement-striving, and perseverance, and 90 openness items spanning introspection, aesthetics, and abstract-thinking.

\begin{Verbatim}
data("items.gpt5.4.example")
data("embeddings.gpt5.4.example")

dim(items.gpt5.4.example)
#> [1] 180   4

head(items.gpt5.4.example, 3)
#>                type            attribute
#> 1 conscientiousness        self-efficacy
#> 2 conscientiousness        self-efficacy
#> 3 conscientiousness achievement-striving
#>                                                    statement ID
#> 1    I am someone who can handle demanding tasks effectively.  1
#> 2  I am someone who feels capable of meeting difficult goals.  2
#> 3 I am someone who sets high standards for my accomplishments. 3

table(items.gpt5.4.example$type, items.gpt5.4.example$attribute)
#>                     abstract-thinking achievement-striving aesthetics
#>   conscientiousness                 0                   30          0
#>   openness                         30                    0         30
#>                     introspection perseverance self-efficacy
#>   conscientiousness             0           30            30
#>   openness                     30            0             0
\end{Verbatim}

Notice the structure of the item pool. The \texttt{GENIE()} function \emph{requires} a data frame with four columns: \texttt{statement} (the item text), \texttt{attribute} (the sub-facet or characteristic the item targets), \texttt{type} (the overarching construct), and \texttt{ID} (a unique identifier). The structure of this \texttt{data.\allowbreak{}frame} is identical to the data frame that is returned by \texttt{AIGENIE()} when the \texttt{items.\allowbreak{}only} flag is set to \texttt{TRUE}.

The package also includes the items' corresponding embedding matrix:

\begin{Verbatim}
dim(embeddings.gpt5.4.example)
#> [1] 1536  180

# Column names are the item IDs
all(colnames(embeddings.gpt5.4.example) %in% as.character(items.gpt5.4.example$ID))
#> [1] TRUE
\end{Verbatim}

Notice that the matrix is \emph{1,536 rows by 180 columns}: the \texttt{GENIE()} function expects embedding \emph{dimensions} in the rows and \emph{items} in the columns. This mirrors the familiar person-by-item layout of questionnaire data; however, the rows represent embedding dimensions rather than respondents. In a conventional network analysis, the association between two items is computed across respondents; here it is computed across the 1,536 embedding dimensions. Each column is one item's position in semantic space, and the network is estimated from how those columns covary. Note that many external embedding tools return matrices with one row per item, and such output must be transposed (e.g., with \texttt{t()}) before it is passed to \texttt{GENIE()}. Embeddings produced by the package directly will not need to be transposed.

With both objects loaded, the pipeline runs in a single call:

\begin{Verbatim}
gpt5.4.results <- GENIE(
  items = items.gpt5.4.example,
  embedding.matrix = embeddings.gpt5.4.example,
  EGA.model = "glasso",
  EGA.algorithm = "walktrap",
  EGA.uni.method = "louvain",
  uva.cut.off = 0.20,
  run.overall = TRUE,
  plot = TRUE
)
\end{Verbatim}

Because \texttt{embedding.\allowbreak{}matrix} is supplied, no embedding is computed and no API key is needed. The package sets the seed internally because bootEGA resamples the embedding matrix; fixing it makes the stability estimates --- and therefore the final item pool --- reproduce exactly across sessions.

Unless \texttt{silently\ =\ TRUE}, the function reports its progress and finishes by printing a summary of the reduction:

\begin{Verbatim}
#>                            AI-Genie Results
#>                            ----------------
#>
#>                            Conscientiousness Items
#> EGA Model: glasso   Embeddings Used: full   Starting N: 90   Final N: 87
#>              Initial NMI: 100           Final NMI: 100
#>
#>                            Openness Items
#> EGA Model: glasso   Embeddings Used: full   Starting N: 90   Final N: 88
#>              Initial NMI: 95.55          Final NMI: 100
#>
#>                           Overall Sample Results
#> EGA Model: glasso   Embeddings Used: full   Starting N: 180  Final N: 175
#>              Initial NMI: 98.64          Final NMI: 100
\end{Verbatim}

It appears that the reduction here is gentle: only five of 180 items were removed (partly because the pool was generated with adaptive prompting and was already close to its intended structure).

Let's follow the \texttt{openness} items' \texttt{reduction\_\allowbreak{}summary} trajectory:

\begin{Verbatim}
gpt5.4.results$item_type_level$openness$reduction_summary
#>                           stage  N       NMI n_removed_at_stage     delta_NMI
#> 1                       Initial 90 0.9554817                  0  0.0000000000
#> 2 Post-UVA / selected embedding 89 0.9551447                  1 -0.0003369791
#> 3          Post-bootEGA / final 88 1.0000000                  1  0.0445183264
\end{Verbatim}

Ninety items entered, eighty-eight survived, and NMI rose from \texttt{0.955} to \texttt{1.000}. An initial NMI of \texttt{0.955} corresponds to a single item assigned to the wrong community. The middle row shows that removing the redundant item lowered NMI very slightly (\texttt{delta\_\allowbreak{}NMI} of \texttt{-0.0003}). The removed item had been assigned to its correct community, so the one misplaced item made up a slightly larger share of the smaller pool (1 of 89 rather than 1 of 90). The misplaced item was then removed at the bootEGA stage for instability (see below), after which the three \texttt{openness} attributes were recovered perfectly. More generally, NMI is not guaranteed to improve at every individual stage, nor is it guaranteed to improve overall; however, and a small dip is unremarkable. A major decline in NMI (at least 3\%) may signal a poorer reduction run.

We can examine how the scale looks now:

\begin{Verbatim}
head(gpt5.4.results$item_type_level$openness$final_items, 4)
#>    ID                                                                statement
#> 1 100    I am someone who lingers over experiences that strike me as beautiful.
#> 2 101     I am someone who likes to connect specific facts to broader concepts.
#> 3 102 I am someone who enjoy exploring questions that cannot be answered by ...
#> 4 103          I am someone who takes time to sort through my private thoughts.
#>           attribute     type EGA_com
#> 1        aesthetics openness       2
#> 2 abstract-thinking openness       3
#> 3 abstract-thinking openness       3
#> 4     introspection openness       1
\end{Verbatim}

The \texttt{attribute} shown above is the label the researcher assigned and \texttt{EGA\_\allowbreak{}com} is the community the network recovered. Here they correspond one-to-one (i.e., introspection maps to EGA community 1, aesthetics to 2, abstract-thinking to 3) which is what an NMI of \texttt{1.000} means in concrete terms.

We can also examine why each of the five items was removed, since every removal is documented:

\begin{Verbatim}
dim(gpt5.4.results$filtering_audit)
#> [1]  5 25

table(gpt5.4.results$filtering_audit$removal_stage)
#> bootEGA     UVA
#>       1       4
\end{Verbatim}

Three of the four redundancy removals came from the conscientiousness pool. Subsetting the audit to those rows shows the formal criterion and its evidence for each:

\begin{Verbatim}
audit <- gpt5.4.results$filtering_audit
uva_rows <- audit[audit$removal_stage == "UVA" &
                    audit$type == "conscientiousness", ]

uva_rows[, c("ID", "attribute", "reason", "diagnostic_value",
             "cutoff", "uva_sweep", "redundant_with_ID")]
#>   ID            attribute                          reason
#> 1  4 achievement-striving Redundancy: wTO = 0.280 >= 0.20
#> 2 28 achievement-striving Redundancy: wTO = 0.213 >= 0.20
#> 3 58 achievement-striving Redundancy: wTO = 0.215 >= 0.20
#>   diagnostic_value cutoff uva_sweep redundant_with_ID
#> 1        0.2797578    0.2         1                46
#> 2        0.2126657    0.2         1                75
#> 3        0.2154339    0.2         1                 9
\end{Verbatim}

Item 4, for instance, was removed in the first UVA sweep because its weighted topological overlap with item 46 was \texttt{0.280}, above the \texttt{0.20} cutoff. The \texttt{reason} column states each decision in a form that can be reported directly.

And the single stability removal:

\begin{Verbatim}
#>   ID     type         attribute removal_stage
#> 5 95 openness abstract-thinking       bootEGA
#>                                       reason item_stability stability_deficit
#> 5 Instability: item stability = 0.420 < 0.75           0.42              0.33
#>   pre_reduction_EGA_community pre_reduction_loading_gap
#> 5                           3                  0.024919
\end{Verbatim}

The \emph{formal} criterion is \texttt{item\_\allowbreak{}stability} of \texttt{0.42} against a cutoff of \texttt{0.75}. Thus, the item landed in the same community in fewer than half of the bootstrap samples. The \emph{diagnostic} columns explain that its \texttt{pre\_\allowbreak{}reduction\_\allowbreak{}loading\_\allowbreak{}gap} was \texttt{0.025}, meaning it loaded almost equally on its assigned community and on a competing one. An item poised between two dimensions is exactly the kind that moves between them under resampling.

Because \texttt{run.\allowbreak{}overall\ =\ TRUE} was set, a pooled analysis across both item types is available:

\begin{Verbatim}
gpt5.4.results$overall$reduction_summary
#>                        stage   N       NMI n_removed_at_stage  delta_NMI
#> 1             Initial pooled 180 0.9863535                  0 0.00000000
#> 2 Post-type-reduction pooled 175 1.0000000                  5 0.01364655

gpt5.4.results$overall$network_plot
\end{Verbatim}

\begin{figure}[!htbp]
    \centering
    \includegraphics[width=0.95\linewidth]{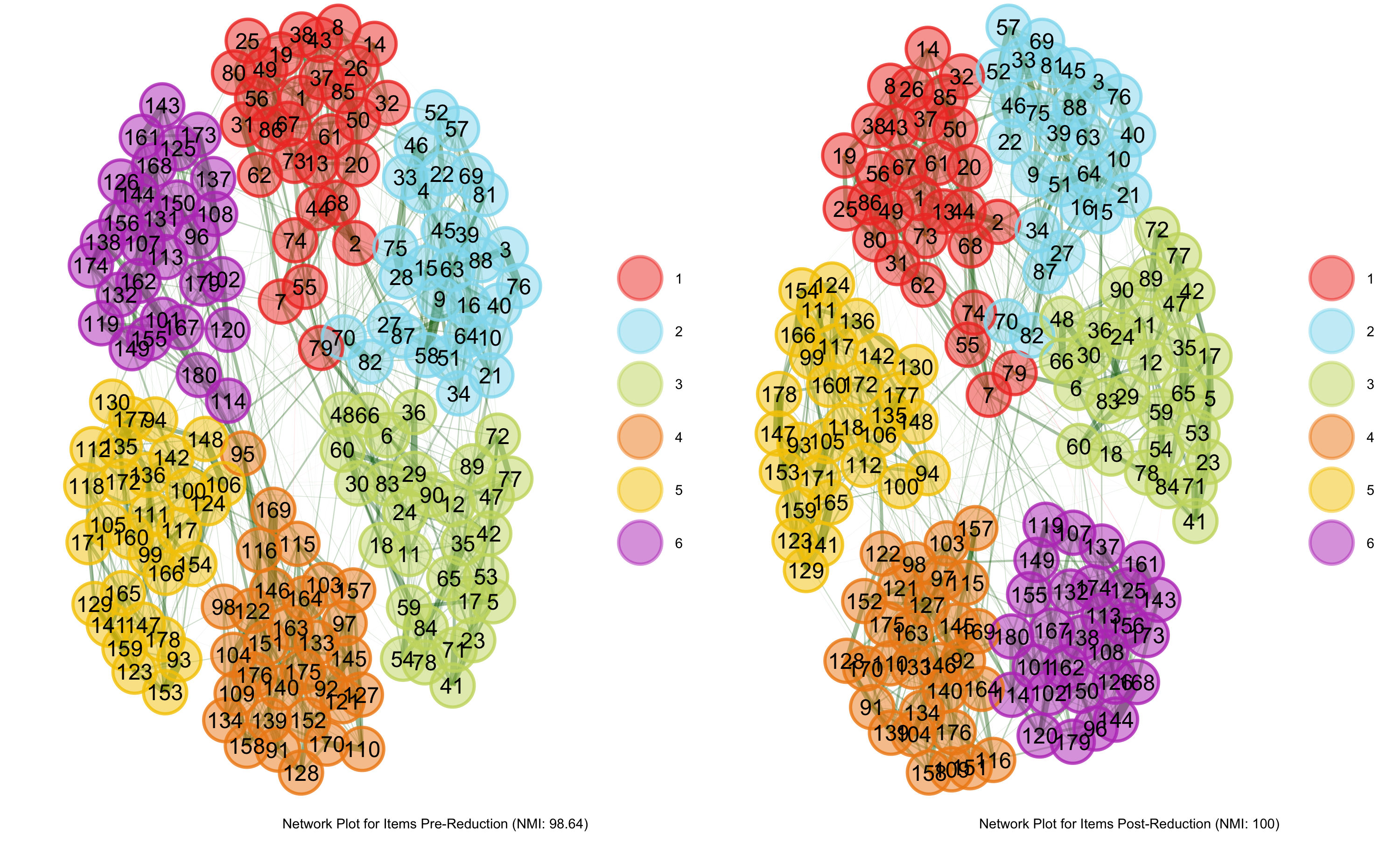}
    \caption{\textit{Bundled GPT-5.4 Example: Pooled EGA Network Before and After Reduction.} The pooled item network before (left) and after (right) type-level reduction for the bundled example. Node color indicates the community assigned by EGA; the six communities correspond to the three attributes within each of the two item types. NMI rises from 98.64 to 100 once the five redundant and unstable items are removed.}
    \label{fig:fig-pooled-network}
\end{figure}

The pooled network in Figure \ref{fig:fig-pooled-network} recovers all six communities (i.e., three attributes within each of the two item types) and reaches an NMI of \texttt{1.000} after reduction.

\subsection{Preparing Your Own Items}\label{preparing-your-own-items}

With the expected shape established, assembling your own pool is a matter of building a large enough \texttt{data.\allowbreak{}frame} object with the same four columns.

The example below prepares 20 hand-written items for an AI Anxiety scale, following the four-factor structure of \textcite{wang2022development} used earlier:

\begin{Verbatim}
my_ai_anxiety_items <- data.frame(
  statement = c(
    # Learning anxiety items
    "I feel overwhelmed by how quickly AI technology is advancing",
    "I worry that my knowledge is not enough to keep up with AI",
    "The complexity of AI systems makes me feel inadequate",
    "I am intimidated by the amount I would need to learn about AI",
    "I feel daunted when I try to understand how AI works",
    # Job replacement items
    "I worry that AI will make my job obsolete",
    "The thought of AI replacing human workers makes me anxious",
    "I am concerned that AI will take over my career field",
    "I feel insecure about my professional future because of AI",
    "I fear that my skills will become irrelevant as AI improves",
    # Sociotechnical blindness items
    "I feel powerless in the face of AI-driven decisions that affect me",
    "I worry about becoming too dependent on AI technology",
    "It bothers me that AI can track and predict my behavior",
    "I worry about losing my privacy to AI-powered surveillance",
    "I am concerned about how much autonomy I am giving up to AI",
    # AI configuration items
    "I do not trust AI systems to make important decisions",
    "I am uncertain about how AI systems actually arrive at their answers",
    "It concerns me that I cannot verify whether AI output is reliable",
    "I feel vulnerable when I have to rely on AI I do not understand",
    "I doubt that AI systems are safe enough to be widely deployed"
  ),
  attribute = c(
    "overwhelmed", rep("inadequacy", 2), rep("intimidated", 2),
    rep("threatened", 2), "replaceable", "insecure", "replaceable",
    "powerless", "overly dependent", rep("surveilled", 2), "powerless",
    "distrustful", rep("uncertain", 2), "vulnerable", "distrustful"
  ),
  type = c(
    rep("learning_anxiety", 5),
    rep("job_replacement", 5),
    rep("sociotechnical_blindness", 5),
    rep("ai_configuration", 5)
  ),
  ID = paste0("AI_", 1:20),
  stringsAsFactors = FALSE
)

# Confirm the shape before running the pipeline
str(my_ai_anxiety_items)
table(my_ai_anxiety_items$type)
\end{Verbatim}

Note that this pool is deliberately small for illustration. With only five items per type it is well below what the bundled example provides. We recommend at \emph{least} 15 items per \texttt{attribute}.

\subsection{Using GENIE with No Attribute/Type Distinction}\label{using-genie-with-no-attributetype-distinction}

It is possible to use the \texttt{GENIE} function if there is only one level of item categorization (i.e., items can only be categorized into distinct types and \emph{not} attributes). However, the \texttt{GENIE} function still expects the item data frame to have the \texttt{type} \emph{and} \texttt{attribute} columns. In this case, the \texttt{attribute} column can be set to any dummy string.

For instance, assume that the \texttt{my\_\allowbreak{}ai\_\allowbreak{}anxiety\_\allowbreak{}items} data frame created in the previous code chunk had no meaningful item attributes. Here is how it would be created instead:

\begin{Verbatim}
# create your data frame the same way as before
no_attributes_df <- data.frame(
  statement = c(
    "I feel overwhelmed by how quickly AI technology is advancing",
    "I worry that my knowledge is not enough to keep up with AI",
    ... # include the item content for ALL items
  ),
  # this column is required, even if it has no meaning.
  attribute = rep("dummyAttribute",  # could be any dummy string
                  20), # twenty items total; same string is used for all items
  type = c(
    rep("learning_anxiety", 5),
    rep("job_replacement", 5),
    ... # include the item types for ALL items
  ),
  ID = paste0("AI_", 1:20),
  stringsAsFactors = FALSE
)
\end{Verbatim}

Because items are \emph{only} categorized into meaningful item types, the default, type-by-type reduction would not be appropriate. Therefore, all items \emph{must} run through the pipeline together. This aggregation can be specified by setting the \texttt{all.\allowbreak{}together} flag to \texttt{TRUE}:

\begin{Verbatim}
no_attributes_res <- GENIE(
  items = no_attributes_df, # the `attribute` column is a dummy string
  openai.API = "your-openai-key",
  embedding.model = "text-embedding-3-small",
  all.together = TRUE # REQUIRED when there are no meaningful item attributes
)
\end{Verbatim}

\subsection{Running the GENIE Pipeline}\label{running-the-genie-pipeline}

When items are supplied without an embedding matrix, \texttt{GENIE()} embeds them first, which requires an API key and an embedding model:

\begin{Verbatim}
genie_results <- GENIE(
  items = my_ai_anxiety_items,
  openai.API = "your-openai-key",
  embedding.model = "text-embedding-3-small",
  plot = TRUE
)

# With four item types, results are organized per-factor
for (factor in names(genie_results$item_type_level)) {
  res <- genie_results$item_type_level[[factor]]
  cat(factor, "- Items retained:", res$final_N, "of", res$start_N,
      "| NMI:", res$initial_NMI, "->", res$final_NMI, "\n")
}

# The complete final validated item pool
do.call(rbind, lapply(genie_results$item_type_level,
                      function(x) x$final_items))
\end{Verbatim}

\subsection{Supplying Pre-Computed Embeddings}\label{supplying-pre-computed-embeddings}

Items that have already been embedded (as, for instance, in a previous session) can be passed directly, skipping the embedding step entirely. The \texttt{embeddings.\allowbreak{}gpt5.4.example} is itself an instance of such a matrix, and shows the required format: embedding dimensions in \emph{rows}, items in \emph{columns}, and column names matching the \texttt{ID} column of the items data frame.

\begin{Verbatim}
genie_results_precomputed <- GENIE(
  items = my_ai_anxiety_items,
  embedding.matrix = my_embeddings,
  plot = TRUE
)
\end{Verbatim}

If the column names and IDs do not correspond, the pipeline cannot align the two objects and will stop with an error rather than silently mismatching items.

Supplying embeddings directly is particularly useful for comparing how different embedding models affect the recovered structure. Because the reduction pipeline is deterministic, any difference in the result is attributable to the embeddings alone.

\section{Part 7: Advanced Prompt Engineering}\label{part7}

The quality of AI-generated items depends heavily on how the LLM is prompted. Recent simulation work within the AI-GENIE framework has demonstrated that prompt engineering strategies can substantially influence the structural validity and redundancy of generated item pools, with effects that scale with model capability \parencite{russelllasalandra2025prompt}. This section covers the advanced prompt customization options available in the \texttt{AIGENIE} package.

\subsection{The Default Prompt Architecture}\label{the-default-prompt-architecture}

When no custom prompts are provided, \texttt{AIGENIE()} automatically constructs prompts using the information supplied through its parameters. The default prompt architecture incorporates several prompt engineering best practices:

\begin{enumerate}
\def\labelenumi{\arabic{enumi}.}
\tightlist
\item
  \textbf{System role (persona prompting):} A domain-specific expert persona is assigned to the model, built from the \texttt{domain}, \texttt{scale.\allowbreak{}title}, and \texttt{audience} parameters.
\item
  \textbf{Contextual instructions:} The prompt includes the construct definition (from \texttt{item.\allowbreak{}type.\allowbreak{}definitions} and/or \texttt{item.\allowbreak{}attribute.\allowbreak{}definitions}), the target attributes, and the response format.
\item
  \textbf{Few-shot examples:} If \texttt{item.\allowbreak{}examples} are provided, they are incorporated to anchor the model's output style.
\item
  \textbf{Adaptive generation:} When \texttt{adaptive\ =\ TRUE} (the default), previously generated items are appended to the prompt to prevent repetition.
\end{enumerate}

\subsection{Customizing the System Role}\label{customizing-the-system-role}

The system role shapes the LLM's overall behavior, and the evidence on whether it helps is worth setting out before showing how to change it. Research on persona prompting has found that assigning a role can improve the quality and coherence of a model's output, particularly for tasks that benefit from domain-specific framing \parencite{russelllasalandra2025prompt,depaoli2023,jiang2024}. The findings are not uniform, however. \textcite{zheng2024} found that personas did not reliably improve performance on factual knowledge tasks, and \textcite{hu2026expert} showed that expert personas can actively \emph{damage} accuracy on knowledge-retrieval tasks while improving alignment-dependent tasks such as writing.

This distinction matters for scale development. Item generation is a creative, domain-specific writing task --- precisely the category in which persona prompting has been shown to help, rather than the knowledge-retrieval category in which it can hurt. The simulation results in \textcite{russelllasalandra2025prompt} confirmed that adding a system role improved item quality within the AI-GENIE pipeline.

A well-crafted system role therefore leads to more domain-appropriate items:

\begin{Verbatim}
# Custom system role
custom_role <- paste(
  "You are an expert psychometrician and clinical psychologist specializing",
  "in technology-related anxiety disorders. You have extensive experience",
  "developing Likert-type self-report measures for adult populations.",
  "Your items should be clear, concise, and written at an 8th-grade reading level.",
  "Each item should be a single complete sentence in first person."
)

results <- AIGENIE(
  item.attributes = ai_anxiety_attributes,
  openai.API = "your-key",
  model = "OpenAI/gpt-5.4",
  system.role = custom_role,
  items.only = TRUE
)
\end{Verbatim}

\subsection{Providing Example Items}\label{providing-example-items}

Example items serve as style and quality anchors. They show the model what ``good'' items look like without directly being copied:

\begin{Verbatim}
# Create example items (format: statement, attribute, type)
# One example per factor to anchor the model's style
example_items <- data.frame(
  statement = c(
    "I feel overwhelmed by how rapidly AI technology is changing",
    "I worry that AI systems will eventually replace my role at work",
    "I feel powerless when AI makes decisions that affect my life",
    "I do not trust AI systems to make fair or accurate judgments"
  ),
  attribute = c("overwhelmed", "threatened", "powerless", "distrustful"),
  type = c("learning_anxiety", "job_replacement",
           "sociotechnical_blindness", "ai_configuration")
)

results <- AIGENIE(
  item.attributes = ai_anxiety_attributes,
  openai.API = "your-key",
  model = "OpenAI/gpt-5.4",
  item.examples = example_items,
  items.only = TRUE
)
\end{Verbatim}

\subsection{Adding Prompt Notes}\label{adding-prompt-notes}

The \texttt{prompt.\allowbreak{}notes} parameter allows users to append brief custom instructions to the automatically constructed prompts. This is useful for adding specific constraints without writing entirely custom prompts:

\begin{Verbatim}
# Global note applied to all item types
results <- AIGENIE(
  item.attributes = ai_anxiety_attributes,
  openai.API = "your-key",
  model = "OpenAI/gpt-5.4",
  prompt.notes = "All items MUST begin with 'I' and should be in present tense. Do not 
  use double-barreled items.",
  items.only = TRUE
)

# Different notes for different item types (for multi-trait scales)
results <- AIGENIE(
  item.attributes = list(
    openness = c("creative", "perceptual", "curious", "philosophical"),
    neuroticism = c("anxious", "depressed", "insecure", "emotional")
  ),
  openai.API = "your-key",
  model = "OpenAI/gpt-5.4",
  prompt.notes = list(
    openness = "Focus on behavioral indicators rather than self-evaluations.",
    neuroticism = "Include both positively and negatively worded items."
  ),
  items.only = TRUE
)
\end{Verbatim}

\subsection{Fully Custom Prompts}\label{fully-custom-prompts}

For maximum control, users can bypass the automatic prompt construction entirely and supply their own prompts:

\begin{Verbatim}
# Fully custom prompts -- one per item type, matching names(ai_anxiety_attributes)
custom_prompts <- list(
  learning_anxiety = paste(
    "Generate novel, high-quality items for a psychological scale measuring",
    "learning anxiety related to AI. Items should capture feelings of being",
    "overwhelmed, inadequacy, or intimidated by AI complexity and advancement.",
    "\n\nGenerate EXACTLY TWO items per attribute. Each item should be a single,",
    "concise, first-person statement suitable for a 6-point Likert response scale.",
    "Items should be distinct from one another and avoid double-barreled phrasing.",
    "\n\nIMPORTANT: Do NOT mix attributes within a single item."
  ),
  job_replacement = paste(
    "Generate novel, high-quality items for a psychological scale measuring",
    "job replacement anxiety related to AI. Items should capture feelings of being",
    "threatened, replaceable, or insecure about one's professional future due to AI.",
    "\n\nGenerate EXACTLY TWO items per attribute. Each item should be a single,",
    "concise, first-person statement suitable for a 6-point Likert response scale.",
    "Items should be distinct from one another and avoid double-barreled phrasing.",
    "\n\nIMPORTANT: Do NOT mix attributes within a single item."
  ),
  sociotechnical_blindness = paste(
    "Generate novel, high-quality items for a psychological scale measuring",
    "sociotechnical blindness anxiety. Items should capture feelings of being",
    "powerless, overly dependent on AI, or surveilled by AI-powered systems.",
    "\n\nGenerate EXACTLY TWO items per attribute. Each item should be a single,",
    "concise, first-person statement suitable for a 6-point Likert response scale.",
    "Items should be distinct from one another and avoid double-barreled phrasing.",
    "\n\nIMPORTANT: Do NOT mix attributes within a single item."
  ),
  ai_configuration = paste(
    "Generate novel, high-quality items for a psychological scale measuring",
    "AI configuration anxiety. Items should capture feelings of distrust,",
    "uncertainty, or vulnerability regarding how AI systems operate and decide.",
    "\n\nGenerate EXACTLY TWO items per attribute. Each item should be a single,",
    "concise, first-person statement suitable for a 6-point Likert response scale.",
    "Items should be distinct from one another and avoid double-barreled phrasing.",
    "\n\nIMPORTANT: Do NOT mix attributes within a single item."
  )
)

results <- AIGENIE(
  item.attributes = ai_anxiety_attributes,
  openai.API = "your-key",
  model = "OpenAI/gpt-5.4",
  main.prompts = custom_prompts,
  items.only = TRUE
)
\end{Verbatim}

Note that when using \texttt{main.\allowbreak{}prompts}, the \texttt{names()} of the list must exactly match the \texttt{names()} of \texttt{item.\allowbreak{}attributes}, and each prompt must mention all attributes for its item type. The package adds formatting instructions automatically to ensure proper parsing of the model's output.

\subsection{Adaptive vs.~Non-Adaptive Generation}\label{adaptive-vs.-non-adaptive-generation}

The \texttt{adaptive} parameter controls whether previously generated items are fed back into the prompt to reduce redundancy. Simulation results show that adaptive prompting is the single most impactful prompt engineering strategy, particularly when paired with newer, higher-capacity models \parencite{russelllasalandra2025prompt}:

\begin{Verbatim}
# Adaptive mode (default and recommended)
results_adaptive <- AIGENIE(
  item.attributes = ai_anxiety_attributes,
  openai.API = "your-key",
  model = "OpenAI/gpt-5.4",
  adaptive = TRUE,  # Previously generated items fed back to prevent repetition
  items.only = TRUE
)

# Non-adaptive mode (for comparison or when context limits are a concern)
results_nonadaptive <- AIGENIE(
  item.attributes = ai_anxiety_attributes,
  openai.API = "your-key",
  model = "OpenAI/gpt-5.4",
  adaptive = FALSE,
  items.only = TRUE
)
\end{Verbatim}

In the simulation study, adaptive prompting reduced item pool redundancy by as much as 93\% for the most capable models, dramatically reducing the number of items flagged by UVA and resulting in larger final item pools with higher structural validity. The effect was most pronounced for newer models (e.g., GPT-5.1, GPT-OSS-120B), with smaller benefits for older or less capable models.

\section{Part 8: Running AI-GENIE Fully Locally}\label{part8}

A distinguishing feature of the \texttt{AIGENIE} package is its support for fully local operation. Using the \texttt{local\_\allowbreak{}AIGENIE()} and \texttt{local\_\allowbreak{}GENIE()} functions, researchers can generate items, create embeddings, and run the entire psychometric pipeline without sending any data to external servers. However, this feature is \emph{not} recommended for most researchers, as running local models requires significant computing power and API servers enable quick use of frontier models.

\subsection{Downloading a Local Model}\label{downloading-a-local-model-1}

Before running the local pipeline, install local model support and download a GGUF model as described in Section \ref{local-setup}. The examples below assume that \texttt{model\_\allowbreak{}path} holds the path returned by \texttt{get\_\allowbreak{}local\_\allowbreak{}llm()} for \emph{Mistral 7B Instruct} (Q4\_K\_M quantization, roughly 4.4 GB).

\subsection{Running the Full Local Pipeline}\label{running-the-full-local-pipeline}

With a model downloaded, the \texttt{local\_\allowbreak{}AIGENIE()} function runs the complete pipeline locally:

\begin{Verbatim}
# Define the construct (same four-factor structure as above)
ai_anxiety_attributes <- list(
  learning_anxiety = c("overwhelmed", "inadequacy", "intimidated"),
  job_replacement = c("threatened", "replaceable", "insecure"),
  sociotechnical_blindness = c("powerless", "overly dependent", "surveilled"),
  ai_configuration = c("distrustful", "uncertain", "vulnerable")
)

# Run fully local AI-GENIE
local_results <- local_AIGENIE(
  item.attributes = ai_anxiety_attributes,
  model.path = model_path,

  # Local embedding model (runs on your machine)
  embedding.model = "sentence-transformers/all-MiniLM-L6-v2",

  # Generation parameters
  target.N = 60,
  adaptive = TRUE,

  # Local inference settings
  n.ctx = 4096, # Context window size
  n.gpu.layers = -1, # Use all GPU layers (set to 0 for CPU-only)
  max.tokens = 1024, # Maximum tokens per generation

  # Embedding settings
  device = "auto",  # Automatically detects GPU
  batch.size = 32,

  # Output settings
  plot = TRUE,
  keep.org = TRUE
)

# Results have the same structure as AIGENIE() output
do.call(rbind, lapply(local_results$item_type_level,
                      function(x) x$final_items))
local_results$item_type_level$learning_anxiety$final_NMI
\end{Verbatim}

\subsection{Local Embedding Options}\label{local-embedding-options}

When running locally, several embedding model families are available, each passed to the \texttt{embedding.\allowbreak{}model} argument. HuggingFace BGE models (e.g., \texttt{BAAI/\allowbreak{}bge-base-en-v1.5}) are likely among the highest-performing open-source options available.

The \texttt{local\_\allowbreak{}GENIE()} function can be used when the researcher already has items but no embeddings:

\begin{Verbatim}
genie_local <- local_GENIE(
  items = my_ai_anxiety_items,
  embedding.model = "sentence-transformers/all-MiniLM-L6-v2",
  device = "auto",  # Automatically uses GPU if available
  plot = TRUE
)
\end{Verbatim}

\subsection{Hardware Considerations}\label{hardware-considerations}

Local inference performance depends heavily on available hardware. For item generation, a quantized 7B parameter model (e.g., Q4 quantization) requires at minimum 8 GB of RAM for CPU-only operation, though 16 GB with Apple Silicon or an NVIDIA GPU will provide substantially faster generation. Machines with 32 GB or more can accommodate larger models. For embedding generation with sentence-transformers, CPU processing is generally adequate for item pools of moderate size, while GPU acceleration (via CUDA or Apple Metal) becomes beneficial for larger pools. The \texttt{device\ =\ "auto"} setting automatically selects the best available hardware.

\begin{Verbatim}
# CPU-only operation (conservative, works everywhere)
local_results <- local_AIGENIE(
  ...,
  n.gpu.layers = 0, # Force CPU for LLM
  device = "cpu" # Force CPU for embeddings
)

# Full GPU acceleration (when available)
local_results <- local_AIGENIE(
  ...,
  n.gpu.layers = -1,  # Offload all layers to GPU
  device = "auto" # Auto-detect GPU for embeddings
)
\end{Verbatim}

\section{Discussion and Conclusion}\label{discussion-and-conclusion}

The present tutorial has introduced the \texttt{AIGENIE} R package, a comprehensive tool for automated psychological scale development and structural validation. The package implements the AI-GENIE framework \parencite{russelllasalandra2026aigenie}, integrating LLM-based item generation with network psychometric methods to produce structurally validated item pools efficiently. This R package gives users an easy and practical way to dive into the emerging realm of Generative Psychometrics.

\subsection{Application Scenarios}\label{application-scenarios}

\textcite{kopka2026taigha} developed the Trust in AI-Generated Health Advice (TAIGHA) scale using AI-GENIE and then carried it through conventional validation, and their reported figures track the argument of this tutorial closely. Sixty items were generated against a two-factor trust/distrust specification; the initial pool returned an NMI of 90.6\% with item stability values ranging from 5\% to 100\%, which the authors describe as poor psychometric properties. Automated reduction retained 28 items, raising NMI to 100\% for both subscales and item stability to between 81\% and 100\%. The authors then recruited ten domain experts who rated the retained items for content relevance, thirty lay participants assessed face validity, and 385 UK participants completed the final instrument. Confirmatory factor analysis supported the two-factor structure with excellent fit (CFI = 0.98, TLI = 0.98, SRMR = 0.03).

Two features of that result are worth drawing out. The first is that the structure recovered \emph{in silico} was the structure subsequently confirmed in human response data. The second is that expert review still removed a substantial number of items on the grounds that they were irrelevant or did not fit the underlying theory.

This post-hoc, expert-led item pruning illustrates that \texttt{AIGENIE} does not \emph{remove} the need for expertise; rather, it fundamentally changes where expert effort is best spent. Traditional development requires content experts to draft items, undergo several rounds of revision, and pilot test before psychometric evaluation can begin; a process that can span months and is extremely costly \parencite{fenn2020developing}. The pipeline relocates expert oversight \emph{away} from drafting candidate items and toward reviewing a computationally curated pool, judging whether its recovered structure is substantively defensible, and designing the empirical validation that must follow.

Developing an instrument has historically required immense funding and infrastructure that many research settings lack, with the result that constructs which are culturally specific, newly emerging, or otherwise underserved often go unmeasured. Lowering the cost of a first structurally evaluated item pool makes creating instruments much more feasible (though, again, \texttt{AIGENIE} does \emph{not} reduce the expertise required to validate them).

\subsection{Practical Recommendations}\label{practical-recommendations}

The findings from both the original AI-GENIE simulation \parencite{russelllasalandra2026aigenie} and the prompt engineering simulation \parencite{russelllasalandra2025prompt} provide empirically grounded recommendations that users can apply directly through the package's parameters. In particular, we recommend enabling adaptive prompting (the default), using newer and more capable models when possible, and taking advantage of the package's contextual parameters (e.g., \texttt{domain}, \texttt{scale.\allowbreak{}title}, \texttt{item.\allowbreak{}type.\allowbreak{}definitions}, \texttt{audience}) to enrich the information available to the LLM. When resources permit, researchers should experiment with multiple models and compare the resulting item pools, as model choice can meaningfully influence item style, diversity, and structural quality. Furthermore, the set of items produced by AI-GENIE still requires a quality and relevance review by experts in the domain. Indeed, developers of the TAIGHA kept domain experts in the loop by subjecting AI-GENIE's items to scrutiny and further pruning \parencite{kopka2026taigha}.

Additionally, the current version of \texttt{AIGENIE} is \emph{only} built for self-report type items. The pipeline was \emph{not} designed for items that assess ability (i.e., items that have a correct answer). However, development of a performance-based \texttt{AIGENIE} is currently underway.

\subsection{Limitations and Considerations}\label{limitations-and-considerations}

It should be noted that AI-GENIE answers \emph{one} question in the item validation argument set out by \textcite{Haladyna2013}, and leaves the rest untouched. It provides evidence about internal structure, obtained in silico. The original AI-GENIE paper demonstrated that in silico structural validity closely tracks empirical structural validity for the best-performing models \parencite{russelllasalandra2026aigenie}, but that correspondence should not be assumed in every application, and it is not a substitute for the remaining sources of validity evidence. Thus, researchers should always inspect the generated items for content validity, cultural appropriateness, and alignment with their theoretical framework.

Moreover, the pipeline optimizes for structural validity (i.e., the alignment between network-detected communities and theoretically specified attributes), but it does not assess whether the items are substantively meaningful, free of cultural bias, or suitable for the intended population. In the TAIGHA development study, expert content review removed a substantial share of the items that AI-GENIE had retained \parencite{kopka2026taigha}. The authors conclude that a manual content validation phase remains necessary, and recommend iteratively refining prompts so that generated items align more closely with the guiding theory. Structural soundness and substantive relevance are separate properties, and the pipeline can only assess the former.

Throughout this paper, NMI recurs as the most visible number the pipeline reports. The reported NMIs should be treated as a \emph{diagnostic} indicator, rather than a target metric to maximize at all costs. In fact, the pipeline does \emph{not} maximize it, as a modest decline at an individual stage is unremarkable, and may even accompany a genuinely better pool. For instance, removing redundant items can lower the apparent agreement between the recovered communities and the intended structure while leaving a less redundant and more stable set of items behind. What a \emph{substantial} decline indicates, however, is a question we have not formally studied (partly because this phenomenon has yet to occur frequently in our simulations). There are many potential explanations for a steep decline, but these explanations span the entire chain on which the pipeline depends (e.g., the encoder model could be too poor or maybe even the researcher's own specification of \texttt{item.\allowbreak{}attributes} is problematic). We therefore offer no diagnostic rule here, though the \texttt{filtering\_\allowbreak{}audit} can provide some insight. From this return object, the researcher may be able to judge whether a disappointing result reflects the items, the models, or even the construct specification.

As just alluded to, the quality of generated items and embeddings depends heavily on the LLM used. Newer and more capable models generally produce better results, particularly when combined with advanced prompting strategies \parencite{russelllasalandra2025prompt}. However, the LLM landscape is evolving rapidly. Models that are state-of-the-art at the time of writing will almost certainly be superseded by the time this paper is read. Regardless, poor embeddings or items may lead to poor results. After all, the pipeline output can only ever be as good as its inputs.

\subsection{Conclusion}\label{conclusion}

The most immediate advantage of working within this \texttt{AIGENIE} framework is the substantial reduction in the time and cost required to produce a structurally valid initial item pool that has been checked for redundancy. Traditional scale development typically requires a team of content experts to draft items, multiple rounds of review and revision, and large-scale pilot testing before psychometric evaluation can even begin. AIGENIE compresses much of this early-stage work into a single function call that generates, embeds, and psychometrically evaluates an item pool in minutes. By packaging state-of-the-art prompt engineering, text embedding, and network psychometric methods into an accessible R interface, AIGENIE lowers the barrier to entry for scale construction. This package could lead to a broader and more diverse set of psychological assessments being available, particularly for constructs that are underrepresented by existing instruments.

The latest version of \texttt{AIGENIE} can always be found on CRAN at \url{https://CRAN.R-project.org/package=AIGENIE}, and the full source code is publicly available.

\printbibliography

@article{fenn2020developing,
  title={Development, Validation and Translation of Psychological Tests},
  author={Fenn, Jessy and Tan, Chee-Seng and George, Sanju},
  journal={BJPsych Advances},
  volume={26},
  number={5},
  pages={306--315},
  year={2020},
  doi={10.1192/bja.2020.33},
  publisher={Cambridge University Press}
}

@article{pearson1901lines,
  title={On Lines and Planes of Closest Fit to Systems of Points in Space},
  author={Pearson, Karl},
  journal={The London, Edinburgh, and Dublin Philosophical Magazine and Journal of Science},
  volume={2},
  number={11},
  pages={559--572},
  year={1901}
}

@article{tengler2025exploring,
  title={Exploring the difference and quality of AI-generated versus human-written texts},
  author={Tengler, Karin and Brandhofer, Gerhard},
  journal={Discover Education},
  volume={4},
  number={1},
  pages={113},
  year={2025},
  publisher={Springer}
}

@article{keane2026using,
  title={Using generative AI to enhance psychometric scale development in market research},
  author={Keane, Darsel and McNaughton, Rod B},
  journal={International Journal of Market Research},
  volume={68},
  number={2},
  pages={194--218},
  year={2026},
  publisher={SAGE Publications Sage UK: London, England}
}

@article{shin2025examining,
  title={Examining the efficacy of generative artificial intelligence in item generation: comparative analysis of human-developed and AI-generated reading tests},
  author={Shin, Dongkwang and Kwon, Suh Keong and Lee, Yongsang},
  journal={Education and Information Technologies},
  volume={30},
  number={16},
  pages={23981--24007},
  year={2025},
  publisher={Springer}
}

@misc{lightman2023letsverifystepstep,
      title={Let's Verify Step by Step},
      author={Hunter Lightman and Vineet Kosaraju and Yura Burda and Harri Edwards and Bowen Baker and Teddy Lee and Jan Leike and John Schulman and Ilya Sutskever and Karl Cobbe},
      year={2023},
      eprint={2305.20050},
      archivePrefix={arXiv},
      primaryClass={cs.LG},
      url={https://arxiv.org/abs/2305.20050},
}

@misc{tao2025,
      title={LLMs are Also Effective Embedding Models: An In-depth Overview},
      author={Chongyang Tao and Tao Shen and Shen Gao and Junshuo Zhang and Zhen Li and Kai Hua and Wenpeng Hu and Zhengwei Tao and Shuai Ma},
      year={2025},
      eprint={2412.12591},
      archivePrefix={arXiv},
      primaryClass={cs.CL},
      url={https://arxiv.org/abs/2412.12591},
}

@article{asudani2023impact,
  title={Impact of word embedding models on text analytics in deep learning environment: a review},
  author={Asudani, Deepak Suresh and Nagwani, Naresh Kumar and Singh, Pradeep},
  journal={Artificial intelligence review},
  volume={56},
  number={9},
  pages={10345--10425},
  year={2023},
  publisher={Springer}
}

@article{russelllasalandra2026aigenie,
  title={Generative Psychometrics via {AI-GENIE}: Automatic Item Generation and Validation with Network-Integrated Evaluation},
  author={Russell-Lasalandra, Lara L. and Christensen, Alexander P. and Golino, Hudson F.},
  journal={Behavior Research Methods},
  volume={58},
  number={8},
  pages={217},
  year={2026},
  doi={10.3758/s13428-026-03082-1}
}

@Manual{aigenie2026software,
  title={{AIGENIE}: Automatic Item Generation and Validation via Network-Integrated Evaluation},
  author={Russell-Lasalandra, Lara L. and Christensen, Alexander P. and Golino, Hudson F.},
  year={2026},
  note={R package version 2.1.2},
  url={https://laralee.github.io/AIGENIE/},
  doi={10.32614/CRAN.package.AIGENIE}
}

@article{vaswani2017attention,
  title={Attention is all you need},
  author={Vaswani, Ashish and Shazeer, Noam and Parmar, Niki and Uszkoreit, Jakob and Jones, Llion and Gomez, Aidan N and Kaiser, {\L}ukasz and Polosukhin, Illia},
  journal={Advances in neural information processing systems},
  volume={30},
  year={2017}
}

@misc{chakrabarty2026,
      title={Can Good Writing Be Generative? Expert-Level AI Writing Emerges through Fine-Tuning on High-Quality Books},
      author={Tuhin Chakrabarty and Paramveer S. Dhillon},
      year={2026},
      eprint={2601.18353},
      archivePrefix={arXiv},
      primaryClass={cs.AI},
      url={https://arxiv.org/abs/2601.18353},
}

@article{russelllasalandra2025prompt,
  title={Prompt Engineering for Scale Development in Generative Psychometrics},
  author={Russell-Lasalandra, Lara L. and Golino, Hudson F.},
  year={2025},
  note={Manuscript under review}
}

@article{golino2017exploratory,
  title={Exploratory Graph Analysis: A New Approach for Estimating the Number of Dimensions in Psychological Research},
  author={Golino, Hudson F. and Epskamp, Sacha},
  journal={PLoS ONE},
  volume={12},
  number={6},
  pages={e0174035},
  year={2017},
  doi={10.1371/journal.pone.0174035}
}

@article{christensen2023unique,
  title={Unique Variable Analysis: A Network Psychometrics Method to Detect Local Dependence},
  author={Christensen, Alexander P. and Garrido, Luis Eduardo and Golino, Hudson F.},
  journal={Multivariate Behavioral Research},
  volume={58},
  number={6},
  pages={1165--1182},
  year={2023},
  doi={10.1080/00273171.2023.2194606}
}

@article{christensen2021estimating,
  title={Estimating the Stability of Psychological Dimensions via Bootstrap Exploratory Graph Analysis: A Monte Carlo Simulation and Tutorial},
  author={Christensen, Alexander P. and Golino, Hudson F.},
  journal={Psych},
  volume={3},
  number={3},
  pages={479--500},
  year={2021},
  doi={10.3390/psych3030032}
}

@incollection{john1999big,
  title={The {Big Five} Trait Taxonomy: History, Measurement, and Theoretical Perspectives},
  author={John, Oliver P. and Srivastava, Sanjay},
  booktitle={Handbook of Personality: Theory and Research},
  editor={Pervin, Lawrence A. and John, Oliver P.},
  edition={2nd},
  pages={102--138},
  year={1999},
  publisher={Guilford Press}
}

@article{hommel2022transformer,
  title={Transformer-Based Deep Neural Language Modeling for Construct-Specific Automatic Item Generation},
  author={Hommel, Bjorn E. and Wollang, Franz-Josef M. and Kotova, Vlada and Zacher, Hannes and Schmukle, Stefan C.},
  journal={Psychometrika},
  volume={87},
  number={2},
  pages={749--772},
  year={2022}
}

@article{gotz2024let,
  title={Let the Algorithm Speak: How to Use Neural Networks for Automatic Item Generation in Psychological Scale Development},
  author={G{\"o}tz, Friedrich M. and Maertens, Rakoen and Loomba, Sahil and Van Der Linden, Sander},
  journal={Psychological Methods},
  volume={29},
  number={3},
  pages={494},
  year={2024}
}

@article{martin2025,
  author  = {Martin Kowal, J. and Hurley Bryant, K. and Segall, D. and Kantrowitz, T.},
  title   = {Harnessing Generative AI for Assessment Item Development: Comparing AI-Generated and Human-Authored Items},
  journal = {International Journal of Selection and Assessment},
  year    = {2025},
  volume  = {33},
  number  = {3},
  pages   = {e70021}
}

@article{boateng2018best,
  title={Best Practices for Developing and Validating Scales for Health, Social, and Behavioral Research: A Primer},
  author={Boateng, Godfred O. and Neilands, Torsten B. and Frongillo, Edward A. and Melgar-Qui{\~n}onez, Hugo R. and Young, Sera L.},
  journal={Frontiers in Public Health},
  volume={6},
  pages={149},
  year={2018}
}

@article{clark2019constructing,
  title={Constructing validity: New developments in creating objective measuring instruments.},
  author={Clark, Lee Anna and Watson, David},
  journal={Psychological assessment},
  volume={31},
  number={12},
  pages={1412},
  year={2019},
  publisher={American Psychological Association}
}

@article{danon2005comparing,
  title={Comparing Community Structure Identification},
  author={Danon, Leon and Díaz-Guilera, Albert and Duch, Jordi and Arenas, Alex},
  journal={Journal of Statistical Mechanics: Theory and Experiment},
  volume={2005},
  number={9},
  pages={P09008},
  year={2005},
  doi={10.1088/1742-5468/2005/09/P09008}
}

@article{garrido2026estimating,
  author  = {Garrido, L. and Russell-Lasalandra, L. L. and Rodriguez-Montoya, C. and Golino, H.},
  title   = {Estimating Dimensional Structure in Generative Psychometrics: Comparing PCA and Network Methods Using Large Language Model Item Embeddings},
  year    = {2026},
  month   = {9},
  day     = {4},
  url     = {https://osf.io/preprints/psyarxiv/2s7pw_v2},
  note    = {PsyArXiv preprint}
}

@article{zhang2005general,
  title={A general framework for weighted gene co-expression network analysis},
  author={Zhang, Bin and Horvath, Steve and others},
  journal={Statistical applications in genetics and molecular biology},
  volume={4},
  number={1},
  pages={1128},
  year={2005}
}

@article{wang2022development,
  title={Development and validation of an artificial intelligence anxiety scale: An initial application in predicting motivated learning behavior},
  author={Wang, Yu-Yin and Wang, Yi-Shun},
  journal={Interactive Learning Environments},
  volume={30},
  number={4},
  pages={619--634},
  year={2022},
  publisher={Taylor \& Francis}
}

@manual{ushey2023reticulate,
  title={reticulate: Interface to Python},
  author={Ushey, Kevin and Allaire, JJ and Tang, Yuan},
  year={2023},
  note={R package version 1.39.0},
  url={https://CRAN.R-project.org/package=reticulate}
}

@misc{astral2024uv,
  title={uv: An Extremely Fast Python Package and Project Manager},
  author={{Astral}},
  year={2024},
  url={https://github.com/astral-sh/uv}
}

@misc{openai2024tokens,
  title={What Are Tokens and How to Count Them?},
  author={{OpenAI}},
  year={2024},
  url={https://help.openai.com/en/articles/4936856-what-are-tokens-and-how-to-count-them}
}

@article{depaoli2023,
  title={Can Large Language Models Help Augment English Psycholinguistic Datasets?},
  author={De Paoli, Stefan},
  journal={arXiv preprint arXiv:2307.14176},
  year={2023}
}

@article{jiang2024,
  title={PersonaLLM: Investigating the Ability of Large Language Models to Express Personality Traits},
  author={Jiang, Hang and Zhang, Xiajie and Cao, Xubo and Kabbara, Jad and Roy, Deb},
  journal={Findings of the Association for Computational Linguistics: NAACL 2024},
  year={2024}
}

@inproceedings{zheng2024,
  title={When ``A Helpful Assistant'' Is Not Really Helpful: Personas in System Prompts Do Not Improve Performances of Large Language Models},
  author={Zheng, Mingqian and Pei, Jiaxin and Jurgens, David},
  booktitle={Findings of the Association for Computational Linguistics: EMNLP 2024},
  year={2024}
}

@article{chen2025unleashing,
  title={Unleashing the Potential of Prompt Engineering in Large Language Models: A Comprehensive Review},
  author={Chen, Banghao and Zhang, Zhaofeng and Langréné, Nicolas and Zhu, Shengxin},
  journal={Patterns},
  volume={6},
  number={6},
  pages={101260},
  year={2025},
  publisher={Elsevier}
}

@article{hu2026expert,
  title={Expert Personas Improve LLM Alignment but Damage Accuracy: Bootstrapping Intent-Based Persona Routing with PRISM},
  author={Hu, Zizhao and Rostami, Mohammad and Thomason, Jesse},
  journal={arXiv preprint arXiv:2603.18507},
  year={2026}
}

@article{golino2020investigating,
  author  = {Golino, Hudson F. and Shi, D. and Christensen, Alexander P. and Garrido, L. E. and Nieto, M. D. and Sadana, R. and Thiyagarajan, J. A. and Martinez-Molina, A.},
  title   = {Investigating the Performance of Exploratory Graph Analysis and Traditional Techniques to Identify the Number of Latent Factors: A Simulation and Tutorial},
  journal = {Psychological Methods},
  year    = {2020},
  volume  = {25},
  number  = {3},
  pages   = {292--320},
  month   = {6},
  doi     = {10.1037/met0000255}
}

@article{kopka2026taigha,
  title={The Trust in {AI}-Generated Health Advice ({TAIGHA}) Scale and Short Version ({TAIGHA-S}): Development and Validation Study},
  author={Kopka, Marvin and Majeed, Azeem and Spinelli, Gabriella and El-Osta, Austen and Feufel, Markus},
  journal={PLOS Digital Health},
  volume={5},
  number={7},
  pages={e0001488},
  year={2026},
  doi={10.1371/journal.pdig.0001488}
}

@article{brickman2025llm,
  title={Large Language Models for Psychological Assessment: A Comprehensive Overview},
  author={Brickman, Jocelyn and Gupta, Mehak and Oltmanns, Joshua R.},
  journal={Advances in Methods and Practices in Psychological Science},
  volume={8},
  number={3},
  pages={1--26},
  year={2025},
  doi={10.1177/25152459251343582}
}

@article{hernandez2023aiip,
  title={The {AI-IP}: Minimizing the Guesswork of Personality Scale Item Development Through Artificial Intelligence},
  author={Hernandez, Ivan and Nie, Weiwen},
  journal={Personnel Psychology},
  volume={76},
  number={4},
  pages={1011--1035},
  year={2023},
  doi={10.1111/peps.12543}
}

@article{fyffe2024transforming,
  title={``Transforming'' Personality Scale Development: Illustrating the Potential of State-of-the-Art Natural Language Processing},
  author={Fyffe, Shea and Lee, Philseok and Kaplan, Seth},
  journal={Organizational Research Methods},
  volume={27},
  number={2},
  pages={265--300},
  year={2024},
  doi={10.1177/10944281231155771}
}

@article{hommel2025infer,
  title={Language Models Accurately Infer Correlations Between Psychological Items and Scales from Text Alone},
  author={Hommel, Bj{\"o}rn E. and Arslan, Ruben C.},
  journal={Advances in Methods and Practices in Psychological Science},
  volume={8},
  number={4},
  pages={1--19},
  year={2025},
  doi={10.1177/25152459251377093}
}

@article{kambeitz2025structure,
  title={The Empirical Structure of Psychopathology Is Represented in Large Language Models},
  author={Kambeitz, Joseph and Schiffman, Jason and Kambeitz-Ilankovic, Lana and Mittal, Vijay A. and Ettinger, Ulrich and Vogeley, Kai},
  journal={Nature Mental Health},
  volume={3},
  number={12},
  pages={1482--1492},
  year={2025},
  doi={10.1038/s44220-025-00527-y}
}

@article{fang2022construct,
  title={Evaluating the Construct Validity of Text Embeddings with Application to Survey Questions},
  author={Fang, Qixiang and Nguyen, Dong and Oberski, Daniel L.},
  journal={EPJ Data Science},
  volume={11},
  number={1},
  pages={39},
  year={2022},
  doi={10.1140/epjds/s13688-022-00353-7}
}

@article{kilmen2025shortening,
  title={Shortening Psychological Scales: Semantic Similarity Matters},
  author={Kilmen, Sevilay and Bulut, Okan},
  journal={Educational and Psychological Measurement},
  volume={85},
  number={5},
  pages={910--934},
  year={2025},
  doi={10.1177/00131644251319047}
}

@book{Haladyna2013,
  author    = {Thomas M. Haladyna and Michael C. Rodriguez},
  title     = {Developing and Validating Test Items},
  year      = {2013},
  publisher = {Taylor \& Francis Group},
  doi = {10.4324/9780203850381}
}

\end{document}